%% file: main.tex
\documentclass[runningheads]{llncs}

\usepackage{eccv}

\usepackage{eccvabbrv}

\usepackage{graphicx}
\usepackage{booktabs}
\usepackage{multirow}
\usepackage[accsupp]{axessibility}  
\usepackage{wrapfig}

\usepackage{colortbl}

\usepackage[pagebackref,breaklinks,colorlinks]{hyperref}
\usepackage[ruled,vlined,linesnumbered]{algorithm2e}
\usepackage{setspace}
\usepackage{orcidlink}
\usepackage{CJKutf8}

\begin{document}

\title{Discrete Diffusion Bridges for Spatiotemporally Aligned Image Translation and Generation}

\titlerunning{Discrete Diffusion Bridges}

\newcommand{\orcidauthorA}{0009-0004-1292-5702}
\newcommand{\orcidauthorB}{0000-0002-7516-5008}
\newcommand{\orcidauthorC}{0000-0001-8578-2539}
\newcommand{\orcidauthorD}{0000-0002-8548-861X}
\newcommand{\orcidauthorE}{0000-0002-8039-6679}
\newcommand{\orcidauthorF}{0000-0003-3805-7654}
\newcommand{\orcidauthorG}{0000-0001-8235-7852}

\author{Xing Xie\inst{1,2}\orcidlink{\orcidauthorA} \and
Jiawei Liu\inst{1}\orcidlink{\orcidauthorB} \and
Shijun Zhou\inst{1,2}\orcidlink{\orcidauthorC} \and
Huijie Fan\inst{1}\thanks{Corresponding author.}\orcidlink{\orcidauthorD} \and
Zhi Han\inst{1}\orcidlink{\orcidauthorE} \and
Yandong Tang\inst{1}\orcidlink{\orcidauthorF} \and
Liangqiong Qu\inst{3}$^{\star}$\orcidlink{\orcidauthorG}}

\authorrunning{X. Xie et al.}

\institute{State Key Laboratory of Robotics and Intelligent Systems, Shenyang Institute of Automation, Chinese Academy of Sciences, Shenyang, China \and
University of Chinese Academy of Sciences, Beijing, China \and
School of Computing and Data Science, The University of Hong Kong, Hong Kong\\
\email{\{xiexing, liujiawei, zhoushijun, fanhuijie, hanzhi, ytang\}@sia.cn, liangqqu@hku.hk}}

\maketitle

\begin{abstract}
    We propose Discrete Diffusion Bridges (DDB), a novel framework designed to resolve the fundamental spatiotemporal misalignment of standard discrete diffusion in image translation and generation. By corrupting data into a pure mask state via a random schedule, the conventional forward process induces a twofold misalignment: spatially, this pure-mask destination entirely discards the rich structural priors of the source image; temporally, the random masking order inherently contradicts the ``easy-first, hard-last'' decoding mechanism used during inference. To address this, DDB constructs a direct and efficient trajectory between domains. Spatially, we introduce a hybrid absorption mechanism that redefines the absorbing state to a stochastic mixture of mask and source tokens, effectively injecting source prior as spatial anchors into the latent space. Temporally, we design an information-guided noise schedule that quantifies semantic variation to prioritize the corruption of high-information regions at earlier timesteps. This ensures the model learns to resolve difficult semantic changes using robust context from invariant regions. Extensive experiments validate the versatility and robustness of our framework across diverse generative paradigms. DDB effectively balances edit alignment with structural fidelity across both text-guided semantic manipulation and pure structural image translation, while inherently complementing text-to-image generation and guaranteeing robust high-quality decoding under extremely low sampling steps. Code and models are available at \href{https://github.com/HKU-HealthAI/DDB}{https://github.com/HKU-HealthAI/DDB}.

\end{abstract}

\section{Introduction}
\begin{figure*}[t]
    \centering
    \includegraphics[width=\textwidth]{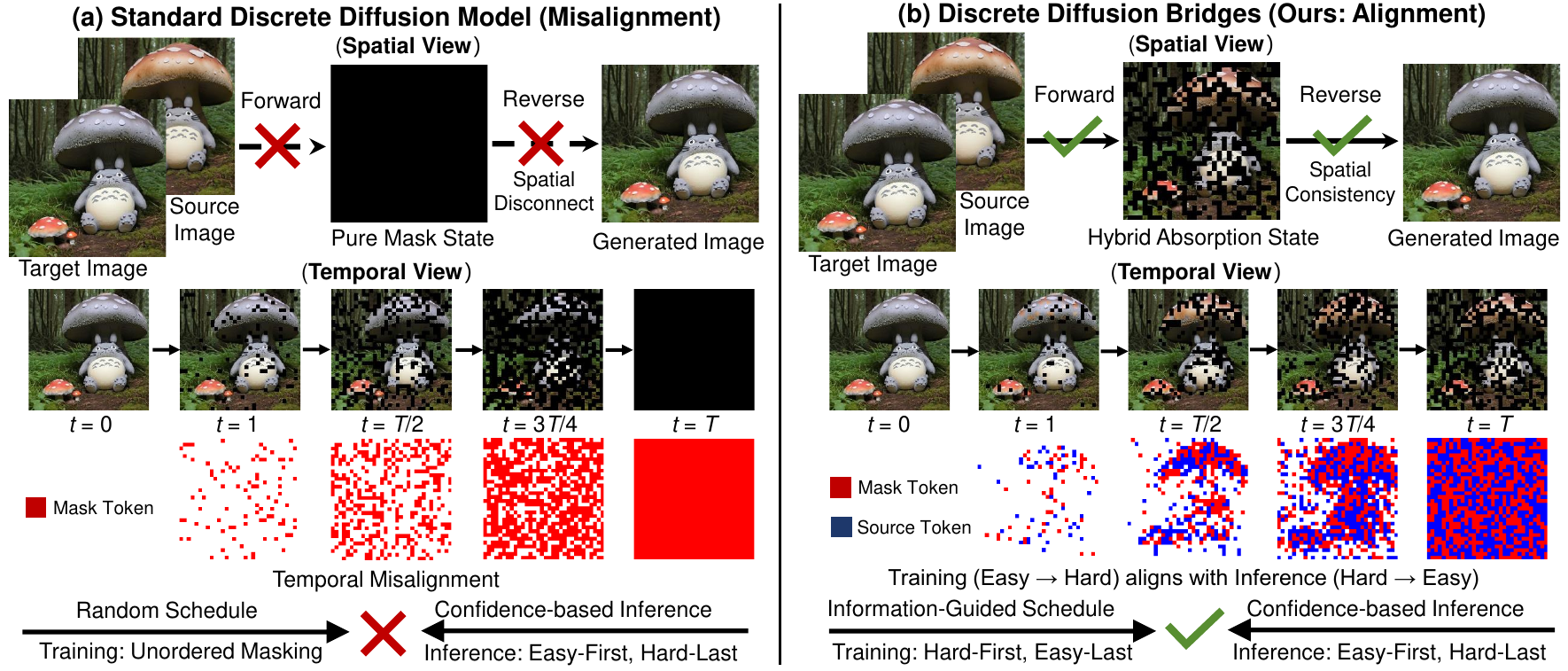}
    \caption{Comparison between Standard Discrete Diffusion and the proposed Discrete Diffusion Bridges (DDB). (a) Standard models suffer from spatiotemporal misalignment: \textbf{spatially} discarding structural priors via a pure mask state, and \textbf{temporally} employing a random training schedule that contradicts the ``easy-first, hard-last'' inference. (b) DDB resolves these discrepancies. \textbf{Spatially}, our Hybrid Absorption mechanism injects source tokens into the mask state as spatial anchors, maintaining structural consistency. \textbf{Temporally}, our Information-Guided Schedule prioritizes masking complex regions early, aligning the training and inference trajectories.}
    \label{fig:first_figure}
\end{figure*}

Discrete diffusion models (DDMs) \cite{sahoo2024simple,ou2024your} have recently emerged as a compelling paradigm in generative AI. Unlike the continuous diffusion models that rely on Gaussian noise corruption, DDMs operate in a compressed latent space using absorbing state formulations \cite{austin2021structured,shi2024simplified}. They employ a discrete masking and prediction paradigm: the forward process absorbs the target data into a pure mask state, while the reverse process progressively recovers the original content, as shown in Fig \ref{fig:first_figure}(a). This discrete nature allows the model to learn the conditional probability of corrupted tokens, enabling high-fidelity generation through iterative refinement. The success of this paradigm has rapidly extended to complex tasks \cite{you2025llada,yang2025mmada}, demonstrating remarkable potential in capturing long-range dependencies and high-frequency details.

However, directly applying the standard discrete diffusion paradigm to image translation and generation tasks reveals a fundamental limitation, which we identify as spatiotemporal misalignment, as shown in Fig \ref{fig:first_figure}(a). Spatially, standard models treat image generation as a ``void-to-image'' reconstruction process. The forward process blindly absorbs the data into a generic, all-mask state (a pure absorbing state), ignoring the explicit structural priors provided by the source image. This forces the model to hallucinate content from scratch rather than transforming the existing semantic layout, leading to a loss of fidelity in identity-preserving tasks. Temporally, a critical discrepancy exists between the training and inference. During training, tokens are masked via a random schedule, treating complex semantic regions (e.g., edited subjects) and simple background regions equally. Conversely, during inference, DDMs typically employ a confidence-based sampling strategy, where easy tokens are determined first and hard tokens last. This contradiction between an unordered training curriculum and a difficulty-aware inference path prevents the model from effectively learning the transformation logic for complex translation tasks.

To bridge these gaps, we propose Discrete Diffusion Bridges, a novel framework that reformulates image translation and generation as a spatiotemporally consistent trajectory. Our core insight is to redefine the ``noise'' in discrete diffusion not as an absence of information, but as a guided transition state, as shown in Fig \ref{fig:first_figure}(b). Specifically, we introduce two synergistic designs. First, to resolve spatial disconnect, we propose hybrid absorption. Instead of driving forward process toward a pure mask, we construct a hybrid absorption space, where the absorbing state is a probabilistic mixture of mask tokens and source tokens. This mechanism effectively injects source image as a spatial anchor directly into diffusion latent space, ensuring generation is grounded in the source context. Second, to resolve temporal disconnect, we devise an information-guided noise schedule. By measuring the translation difficulty between the source and target, we construct a robust stochastic masking curriculum that prioritizes the corruption of high-information regions at earlier timesteps. This ensures that the training process mirrors the ``easy-first, hard-last'' dynamics of inference, forcing model to tackle difficult semantic changes only when sufficient context is available.

Extensive experiments demonstrate that DDB establishes a highly efficient and unified bridge across diverse generative paradigms. Specifically, in Image-to-Image (I2I) translation, DDB achieves a state-of-the-art balance between edit alignment and source context preservation. Whether executing precise semantic manipulation (e.g., image editing, style transfer) or performing structural mapping (e.g., all-in-one restoration, modality translation), DDB consistently delivers high-fidelity outputs. Furthermore, DDB demonstrates highly competitive performance in Text-to-Image (T2I) generation, proving that our spatiotemporal alignment mechanism naturally complements universal generative priors. Overall, these results confirm the broad applicability and robustness of our framework. Driven by the spatiotemporal trajectory, DDB effectively balances generation diversity and structural fidelity, while maintaining excellent quality at extremely low sampling steps.
Our contributions are summarized as follows:

(\romannumeral1) 
We identify the spatiotemporal misalignment in applying discrete diffusion to image translation and generation, and propose Discrete Diffusion Bridges, a unified framework that harmonizes state space and generative trajectory.

(\romannumeral2) 
We introduce Hybrid Absorption, a novel transition mechanism that replaces the standard pure-mask absorbing state with a source-aware hybrid state, robustly preserving spatial structure.

(\romannumeral3) 
We design an Information-Guided Noise Schedule, which decouples the stochastic nature of the noise state from a deterministic, difficulty-aware order, aligning training objectives with inference process.

(\romannumeral4)  
Extensive experiments validate the versatility of DDB across diverse generative paradigms. It effectively balances edit alignment with structural fidelity in I2I translation tasks, inherently complements T2I generation, and maintains excellent quality under limited sampling steps.

\section{Related Work}

\subsection{Discrete Diffusion Modeling}
Discrete diffusion models \cite{austin2021structured,sahoo2024simple} have emerged as a powerful generative paradigm, offering a highly parallelizable alternative to autoregressive models. This approach traces back to masked language modeling in natural language processing, like BERT \cite{devlin2019bert}, which reconstructs obscured tokens using bidirectional context. MaskGIT \cite{chang2022maskgit} and MUSE \cite{chang2023muse} adapted this to the visual domain via iterative token refinement for parallel decoding. Modern masked diffusion models \cite{bai2024meissonic,zhu2025di,rojas2025diffuse} formulate this as a discrete-state Markov chain, corrupting data to an absorbing state and utilizing bidirectional attention for parallel recovery. 
Recently, this paradigm has scaled to billion-parameter language models \cite{nie2025large,ye2025dream,song2025seed}, achieving performance comparable to advanced AR models at the billion-parameter scale.
It has also driven the development of discrete multimodal large language models \cite{you2025llada,yu2025dimple,luo2025reinforcement,bai2025masks,zhu2025soft}. Notably, LaViDa-O \cite{li2025lavida} introduces an Elastic-MoT architecture and stratified sampling to efficiently balance multimodal tasks. Muddit \cite{shi2025muddit} adopts a pure diffusion transformer for high-quality, unified generation.

\subsection{Image Translation and Generation}
Continuous diffusion models \cite{ho2020denoising,rombach2022high,peebles2023scalable} have achieved remarkable success across various image translation \cite{huang2025diffusion,chen2025unirestore,Wang_PSTM,10504912,qu2015pixel} and generation tasks \cite{cao2025controllable,xie2026dvg,huang2026exposurebiasalleviatedirectional,dong2026bring,WM_Survey}. Early models like Palette \cite{saharia2022palette} and SR3 \cite{saharia2022image} condition on input images to map pure noise to clean targets. To enhance cross-domain mapping, subsequent works such as DDBM \cite{zhou2024denoising}, RDDM \cite{liu2024residual}, DDIB \cite{su2022dual} and BBDM \cite{li2023bbdm} establish direct trajectories between source and target domains utilizing residuals and diffusion bridges.
Inspired by large language models, recent research explores the autoregressive paradigm \cite{esser2021taming,ramesh2021zero} for visual tasks.
Several studies \cite{sun2024autoregressive,team2024chameleon,lu2024unified,liu2024lumina,xie2026unleashing} attempt to unify text and image modeling within a single LLM framework, boosting performance via tokenizer optimization \cite{sun2024autoregressive}, early fusion modeling \cite{team2024chameleon}, and flexible resolution adaptation \cite{liu2024lumina}. However, their strict left-to-right decoding intrinsically suffers from slow inference speeds and struggles to maintain global spatial coherence.
To overcome these AR limitations, discrete diffusion models have emerged as a promising solution for unified generation. Frameworks like UniDisc \cite{swerdlow2025unified} and Muddit \cite{shi2025muddit} refine tokens iteratively, enabling rapid, parallel decoding. Lumina-DiMOO \cite{xin2025lumina} leverages structural tokens to natively support arbitrary-resolution generation. 

\begin{figure*}[t]
    \centering
    \includegraphics[width=\textwidth]{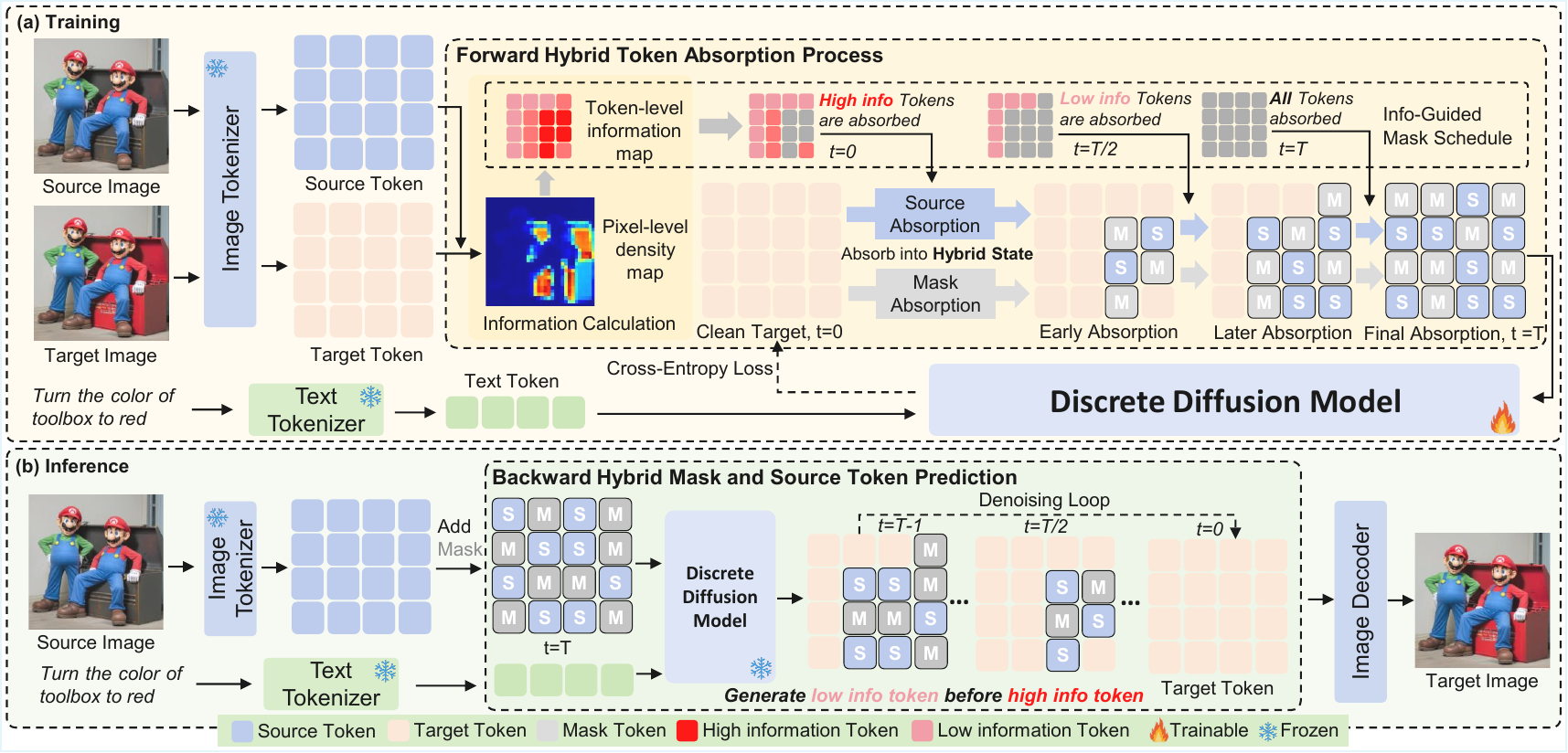}
    \caption{Overview of the Discrete Diffusion Bridges (DDB) framework. (a) Training: The forward process employs Hybrid Absorption to corrupt target tokens into a source-mask mixture, establishing explicit spatial anchors. It employs an Information-Guided Schedule: a pixel-level density map is computed and downsampled into a token-level information map, prioritizing the early absorption of high-information tokens. (b) Inference: Starting from the hybrid state, confidence-based decoding resolves low-information regions before complex edits, achieving spatiotemporal alignment with training process.}
    \label{fig:pipline}
\end{figure*}

\section{Method}

\subsection{Motivation: Misalignment in Standard DDM}
\label{sec:motivation}
We first revisit the standard discrete diffusion models (DDMs) to formalize both their forward and reverse processes, thereby illuminating their inherent limitations in image translation and generation. Let $\mathcal{K} = \{1, \dots, K\}$ be a discrete codebook vocabulary of size $K$. A target image can be tokenized into a discrete sequence $\mathbf{z}_0 \in \mathcal{K}^N$, where $N$ is the sequence length.

\textbf{Forward Process.} Standard DDMs define the forward process $q(\mathbf{z}_t | \mathbf{z}_0)$ as a gradual corruption that progressively replaces input tokens with a special mask token $[\text{M}]$. Once a token is masked, it remains in that state throughout the remainder of the process, making $[\text{M}]$ an absorbing state. At timestep $t \in [0, T]$, the forward transition for the $i$-th token is defined as:
\begin{equation}
\label{eq.forward}
q(z_t^{(i)} | z_0^{(i)}) = \text{Cat}(z_t^{(i)}; (1 - \gamma(t)) z_0^{(i)} + \gamma(t) m).
\end{equation}
Cat(·) denotes a categorical distribution. Where $z_0^{(i)}$ and $m$ are the one-hot vectors corresponding to original target token and $[\text{M}]$ token, respectively. $\gamma(t) \in [0, 1]$ is a monotonically increasing noise schedule such that $\gamma(0) \approx 0$ and $\gamma(T) = 1$. Crucially, this transition probability $\gamma(t)$ is applied uniformly across all spatial locations, meaning the masking process is entirely unordered and random.

\textbf{Reverse Process.} During inference, generation proceeds via an iterative refinement from the pure mask state $\mathbf{z}_T = [\text{M}]^N$ to $\mathbf{z}_0$. Tokens unmasked in previous steps are carried over unchanged. At each reverse step $t \to t - 1$, the network predicts the categorical distribution $p_\theta(\mathbf{z}_0 | \mathbf{z}_t, \mathbf{c})$ given a condition $\mathbf{c}$. For each currently masked token position $i$ (where $z_t^{(i)} = [\text{M}]$), the most probable candidate $\hat{z}_0^{(i)}$ and its corresponding confidence score $s_t^{(i)}$ are computed:
\begin{equation}
\hat{z}_0^{(i)} = \arg\max_{k \in \mathcal{K}} p_\theta(z_0^{(i)}=k | \mathbf{z}_t, \mathbf{c}), \quad s_t^{(i)} = \max_{k \in \mathcal{K}} p_\theta(z_0^{(i)}=k | \mathbf{z}_t, \mathbf{c}).
\end{equation}
The model then ranks the newly predicted candidates by their confidence scores and retains a specific proportion of the most confident tokens as governed by the noise schedule. The reverse transition is thus formulated as:
\begin{equation}
\label{eq.reverse}
z_{t-1}^{(i)} = 
\begin{cases}
z_t^{(i)} & \text{if } z_t^{(i)} \neq [\text{M}] \\
\hat{z}_0^{(i)} & \text{if } z_t^{(i)} = [\text{M}] \text{ and } s_t^{(i)} \ge \tau_t \\
[\text{M}] & \text{otherwise}
\end{cases}
\end{equation}
where $\tau_t$ is a dynamic confidence threshold implicitly determined by the schedule $\gamma(t-1)$ to regulate the decoding pace.

\textbf{Spatiotemporal Misalignment.} When applied to conditional image translation and generation, the discrepancy between Eq.~\ref{eq.forward} and Eq.~\ref{eq.reverse} exposes a fundamental spatiotemporal misalignment. 
\textbf{Spatially}, the forward process strictly drives the data into a pure mask state ($\mathbf{z}_T = [\text{M}]^N$). This completely severs the spatial correspondence between the target and the explicit source priors, forcing the model to hallucinate invariant structures from a semantic void rather than transforming existing source image features. 
\textbf{Temporally}, the generation trajectory in Eq.~\ref{eq.reverse} employs a confidence-driven progressive decoding mechanism. It naturally tends to resolve deterministic features before generating complex and ambiguous semantic details. Conversely, the forward process in Eq.~\ref{eq.forward} applies uniform random masking, forcing the model to learn the corruption of simple and complex regions without any difficulty-aware curriculum. This stark contradiction between random training trajectory and deterministic, confidence-based inference dynamics makes learning the transition mapping highly inefficient.

\subsection{Overview of DDB}
\label{sec:overview}
To overcome these misalignments, we propose \textbf{Discrete Diffusion Bridges} (see Fig. \ref{fig:pipline}), a novel framework that reconceptualizes the generative trajectory. DDB constructs a direct, efficient bridge connecting the target domain to the source domain. The framework comprises two core phases:
(1) A Forward Process (Sec.~\ref{sec:forward}) that builds the bridge. It introduces a Hybrid Absorption mechanism to establish spatial anchors by injecting source priors into the absorbing state, and an Information-Guided Schedule to map difficulty-aware temporal path.
(2) A Reverse Process (Sec.~\ref{sec:reverse}) that efficiently traverses this bridge. Guided by the spatial anchors, it resolves simple invariant regions before complex semantic edits, naturally matching the training trajectory.


\subsection{Forward Process: Constructing the Bridge}
\label{sec:forward}

\subsubsection{Spatial Bridge Anchors: Hybrid Absorption.}
Standard discrete diffusion replaces all corrupted tokens with $[\text{M}]$, which mathematically corresponds to driving the generative state towards a pure mask vector $m$. This mechanism entirely severs the spatial connection between the source and target domains. To construct a direct bridge, we propose a Hybrid Absorption mechanism that injects source priors into the latent space to establish robust spatial anchors.

\textbf{(1) Hybrid State Space.} We define the absorbing state as a source-conditioned hybrid distribution. For a given source image token $y^{(i)}$ represented as a one-hot vector, we define the hybrid absorbing state $h^{(i)}$ as a stochastic mixture of the source prior and the mask:
\begin{equation}
h^{(i)} = \lambda y^{(i)} + (1 - \lambda) m
\end{equation}
where $\lambda \in [0, 1]$ is the source injection ratio. When $\lambda > 0$, the retained source tokens successfully serve as explicit ``spatial anchors'' to bridge the two domains. When $\lambda = 0$, the framework can naturally degrade to a standard discrete diffusion model without other modification.

\textbf{(2) Hybrid Forward Transition.} The forward transition $q(\mathbf{z}_t | \mathbf{z}_0, \mathbf{y})$ is consequently conditioned on both the target and the source. Let $\mathbf{m}_t \in \{0, 1\}^N$ be the binary mask sequence at timestep $t$ (which will be determined by our information-guided schedule), where $m_t^{(i)} = 0$ indicates that the $i$-th token is selected for corruption. The transition for our discrete bridge is formulated as a categorical distribution:
\begin{equation}
q(z_t^{(i)} | z_0^{(i)}, y^{(i)}, m_t^{(i)}) = \text{Cat}\left(z_t^{(i)}; m_t^{(i)} z_0^{(i)} + (1 - m_t^{(i)})h^{(i)} \right)
\end{equation}
Here, $\text{Cat}(\cdot)$ denotes a categorical distribution. This formulation ensures endpoint $\mathbf{z}_T$ retains rich structural priors. 

\vspace{-4mm}
\subsubsection{Temporal Bridge Trajectory: Information-Guided Mask Schedule.}
To explicitly align the training curriculum with inference dynamics, we introduce an Information-Guided Schedule. By quantifying the translation difficulty (information density) of each region, we construct a temporal trajectory that prioritizes corrupting high-information tokens at earlier timesteps.

\textbf{(1) Information Metric Formulation.} We define a token-level information map $\mathcal{I} \in \mathbb{R}^N$ to guide mask schedule, derived from a pixel-level density map $\mathcal{M} \in \mathbb{R}^{H_{px} \times W_{px}}$. 
For translation tasks with source images (e.g., image editing), $\mathcal{M}$ is the absolute pixel-wise difference between the target $\mathbf{X}_0$ and source $\mathbf{X}_{src}$ across $C$ channels, computed as $\mathcal{M} = \frac{1}{C} \sum_{c=1}^{C} |\mathbf{X}_{0, c} - \mathbf{X}_{src, c}|$. 
For generation tasks without source images (e.g., text-to-image), we evaluate structural complexity using the local pixel variance of the grayscale target $\mathbf{X}_{0, g}$ within a $k \times k$ window, formulated as $\mathcal{M} = \text{AvgPool}_{k \times k}(\mathbf{X}_{0, g}^2) - (\text{AvgPool}_{k \times k}(\mathbf{X}_{0, g}))^2$. 
The pixel map $\mathcal{M}$ is then downsampled via average pooling to match the tokenizer's spatial stride, yielding the flattened token-level map $\mathcal{I}$.

\textbf{(2) Robust Stochastic Masking.} 
Strictly sorting tokens by $\mathcal{I}$ yields a deterministic trajectory but restricts exposure to diverse semantic contexts, risking overfitting. To preserve contextual robustness, we propose a stochastic masking strategy. We normalize $\mathcal{I}$ into a probability distribution $P_{info}^{(i)} = \mathcal{I}^{(i)} / \sum_{j=1}^N \mathcal{I}^{(j)}$. The final sampling probability $P_{final}^{(i)}$ for the $i$-th token is formulated as a weighted combination of $P_{info}^{(i)}$ and a uniform distribution $P_{uniform}^{(i)} = 1/N$:
\begin{equation}
P_{final}^{(i)} = (1 - \rho) \cdot P_{info}^{(i)} + \rho \cdot P_{uniform}^{(i)},
\end{equation}
where $\rho \in [0, 1]$ regulates the stochasticity. At timestep $t$, we sample exactly $n_t = \lfloor N \cdot \gamma(t) \rfloor$ mask indices based on $P_{final}$. This multinomial sampling statistically prioritizes high-information regions, maintaining the ``hard-first'' temporal bridge trajectory while intrinsically injecting contextual diversity via $\rho$.

\subsection{Reverse Process: Inference along the Bridge}
\label{sec:reverse}
During inference, the reverse generation precisely traces back along the constructed bridge via an iterative decoding and anchor-resampling strategy. The initial state $\mathbf{z}_T$ is instantiated by independently sampling each token $z_T^{(i)}$ from the hybrid absorbing distribution $h^{(i)}$.

Crucially, tokens that have been confidently resolved in previous steps are carried over unchanged. At each reverse step $t \to t-1$, the network predicts the categorical distribution $p_\theta(\mathbf{z}_0 | \mathbf{z}_t, \mathbf{y}, \mathbf{c})$ given the source anchors $\mathbf{y}$ and condition $\mathbf{c}$. For each currently unresolved token position $i$, we determine the most probable candidate $\hat{z}_0^{(i)}$ and its confidence score $s_t^{(i)}$:
\begin{equation}
\hat{z}_0^{(i)} = \arg\max_{k \in \mathcal{K}} p_\theta(z_0^{(i)}=k | \mathbf{z}_t, \mathbf{y}, \mathbf{c}), \quad s_t^{(i)} = \max_{k \in \mathcal{K}} p_\theta(z_0^{(i)}=k | \mathbf{z}_t, \mathbf{y}, \mathbf{c})
\end{equation}

To align with the training trajectory, the model ranks these newly predicted candidates by their confidence scores. We retain a specific proportion of the most confident tokens as governed by the discrete schedule. Instead of reverting the uncertain regions to a static state or a pure semantic void, we dynamically resample them from the hybrid distribution to perfectly match the stochastic source injection used during training. The transition is formulated as:
\begin{equation}
z_{t-1}^{(i)} = 
\begin{cases}
z_t^{(i)} & \text{if token } i \text{ is already unmasked} \\
\hat{z}_0^{(i)} & \text{if token } i \text{ is newly unmasked } (s_t^{(i)} \ge \tau_{t-1}) \\
\tilde{h}^{(i)} & \text{otherwise} \quad (\text{where } \tilde{h}^{(i)} \sim h^{(i)})
\end{cases}
\end{equation}
where $\tau_{t-1}$ is the dynamic confidence threshold implicitly determined by the schedule $\gamma(t-1)$, and $\tilde{h}^{(i)}$ is a newly drawn sample from the hybrid absorbing distribution $h^{(i)} = \lambda y^{(i)} + (1 - \lambda) m$. 
This ensures undecoded regions receive continuous guidance from dynamic spatial anchors.


\vspace{-2mm}
\subsubsection{Training Objective.}
Let $\mathbf{c}$ be the conditioning signal (e.g., text prompt) and $\mathbf{y}$ be the source image tokens. The model is trained to predict the original target tokens $\mathbf{z}_0$ from the hybrid corrupted state $\mathbf{z}_t$.
We optimize the model using a standard cross-entropy loss applied specifically to the absorbed regions:
\begin{equation}
\mathcal{L}_{DDB} = \mathbb{E}_{t, \mathbf{z}_0, \mathbf{y}, \mathbf{c}} \left[ \sum_{i=1}^N - (1 - m_t^{(i)}) \log p_\theta(z_{0}^{(i)} | \mathbf{z}_t, \mathbf{y}, \mathbf{c}) \right]
\end{equation}

\section{Experiments Analysis and Results}
In the experiments, we comprehensively evaluate the effectiveness of DDB across diverse translation and generation paradigms. First, we validate the superiority of DDB framework in Image-to-Image (I2I) translation across two distinct paradigms: (1) instruction-based translation (image editing, style transfer and sub-driven generation), which validates model's capability for precise \textbf{semantic manipulation} while preserving invariant contexts; and (2) pure image translation (all-in-one restoration, modality translation and super resolution), which verifies its capability for robust \textbf{structural mapping} without textual cues. Next, we show that DDB's advantages extend to unanchored Text-to-Image (T2I) generation, confirming that our spatiotemporal alignment mechanism inherently complements universal \textbf{generative priors}. We then conduct comprehensive component analyses to verify the efficacy of our core designs, including hybrid absorption mechanism and information-guided mask schedule. Finally, we investigate DDB's inference efficiency and robustness under few-step settings.

\subsection{Experimental Setup}
\subsubsection{Datasets.}
For TI2I translation tasks, we use OmniEdit \cite{wei2024omniedit} dataset for image editing and style transfer tasks, and Graph-200K \cite{li2025visualcloze} dataset for subject-driven generation task. For pure I2I translation tasks, we use CDD-11 \cite{guo2024onerestore} dataset for All-in-One restoration task, SynthRAD \cite{thummerer2023synthrad2023} dataset for CT-to-MRI and MRI-to-CT modality translation task, and IXI \cite{ixi_dataset} dataset for super resolution task. For T2I generation, we use MIMIC-CXR \cite{johnson2019mimic} for medical report-to-image generation.

\vspace{-3mm}
\subsubsection{Evaluation Metrics.}
For TI2I tasks, we evaluate background preservation via DINO \cite{oquab2023dinov2} and edit semantic alignment via CLIP-T \cite{radford2021learning}, and overall editing quality using the LLM-based Edit Score \cite{wu2025editreward}.
For pure I2I tasks, we assess pixel-level fidelity using PSNR and SSIM \cite{wang2004image}, and perceptual similarity using LPIPS \cite{zhang2018unreasonable}.
For the T2I task, we use FID \cite{heusel2017gans} and MS-SSIM \cite{wang2003multiscale} to evaluate image fidelity, and CLIP-Score \cite{zhang2023biomedclip} to evaluate text-image alignment.

\vspace{-3mm}
\subsubsection{Implementation Details.}
We adopt Lumina-DiMOO \cite{xin2025lumina} as the backbone of our DDB framework, which is an advanced discrete diffusion model with 8B parameters. This backbone uses VQ-VAE \cite{van2017neural} for latent encoding. Additional training details and all other experimental settings are provided in Appendix.

\subsection{Performance on I2I Translation Tasks}
\label{sec:TI2T task}
To comprehensively evaluate the capability of DDB on I2I translation tasks, we design experiments from two primary perspectives: instruction-based translation (TI2I) for evaluating complex semantic manipulation, and pure image translation (pure I2I) for assessing robust structural mapping. 
\subsubsection{Semantic Manipulation: DDB for TI2I Translation.} 
\begin{figure*}[t]
    \centering
        \centering
        \includegraphics[width=\textwidth]{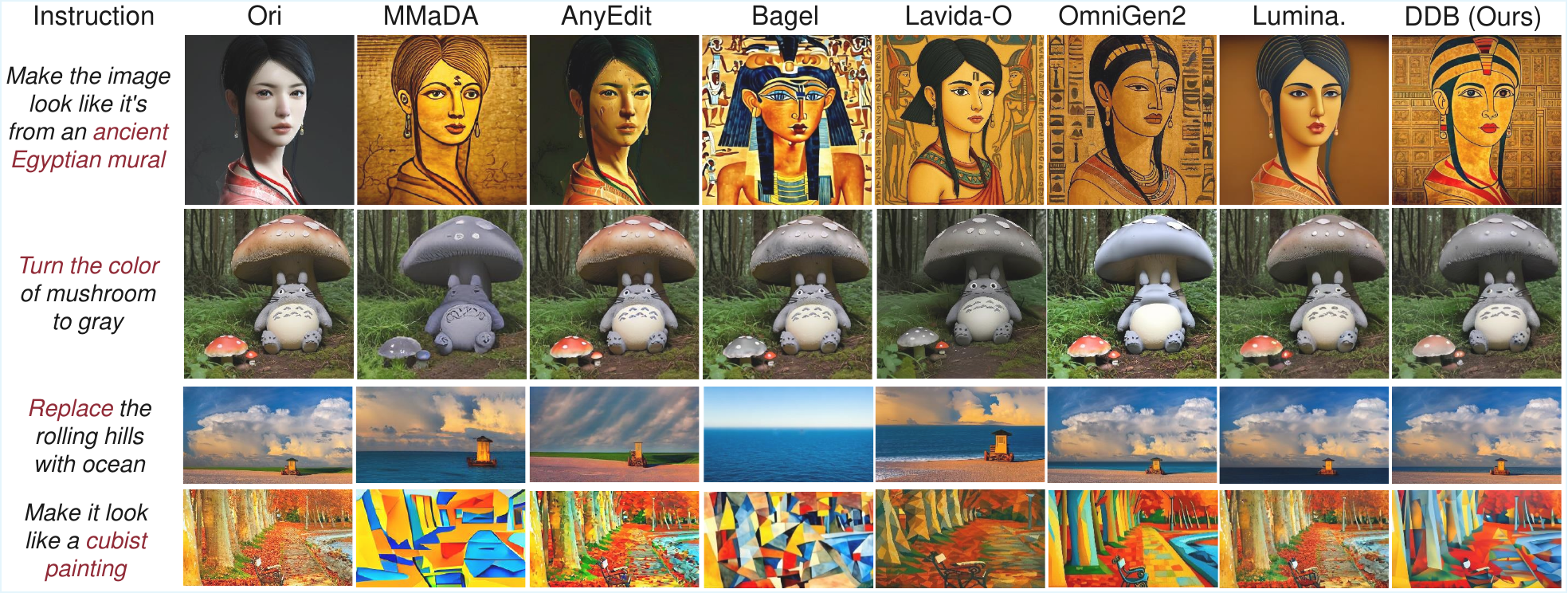}
        \caption{Visual comparison of different methods and DDB for image editing and style transfer tasks on the OmniEdit dataset.}
        \label{fig:edit_style_sub}
\end{figure*}

To assess DDB's capability in text-guided semantic manipulation, we evaluate three distinct sub-tasks: modifying specific regions (Image Editing), globally transforming visual appearance (Style Transfer), and contextualizing background-free target (Subject Driven Generation). As reported in Table \ref{tab:style_subject_editing}, standard discrete diffusion models struggle to balance edit extent (CLIP-T) with source preservation (DINO). In contrast, DDB achieves the highest overall Edit Score by excelling in both metrics. This quantitative superiority reflects two core capabilities of our model: the information-guided trajectory enables precise spatial localization for regional edits, while the hybrid absorption mechanism ensures strict identity retention during aggressive global transformations. Visually (Fig. \ref{fig:edit_style_sub}), DDB strictly preserves unedited background regions and seamlessly blends edited subjects into the scene with high fidelity, mitigating the unintended background alterations commonly observed in baseline methods.

\input{table/edit_omni_bench}

\begin{figure*}[t]
    \centering
        \centering
        \includegraphics[width=\textwidth]{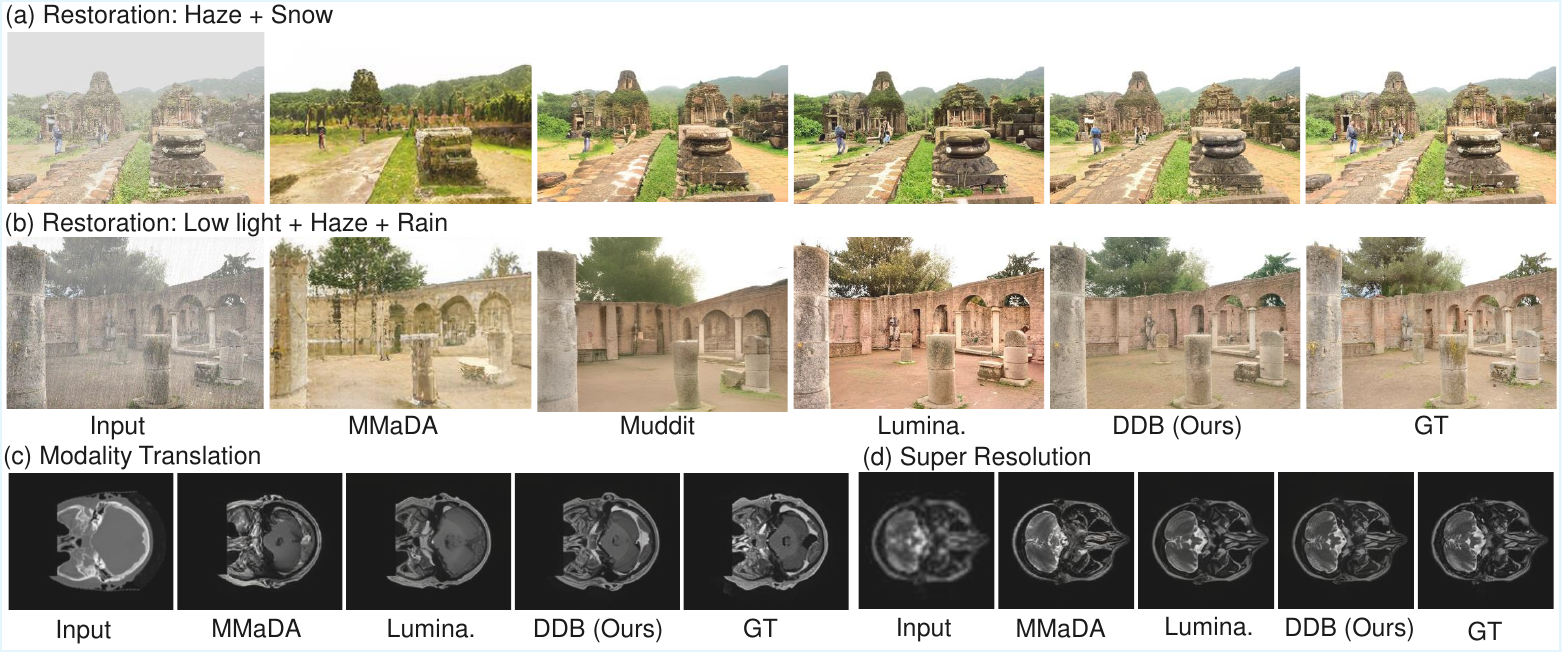}
        \caption{Visual comparison of different methods for all-in-one restoration, modality translation and super resolution tasks on the CDD-11, SynthRAD and IXI datasets.}
        \label{fig:i2i_translation}
\end{figure*}

\input{table/I2I_translation_all}
\subsubsection{Structural Mapping: DDB for Pure I2I Translation.} 
To verify DDB's capacity to capture complex spatial correspondences relying solely on visual priors, we evaluate the model on pure I2I tasks without explicit textual guidance. We deliberately design experiments across three distinct tasks to validate different dimensions of the model's structural mapping capabilities: (1) \textit{All-in-One Restoration} tests the model's multi-task versatility in handling diverse and unpredictable degradations within a single unified framework; (2) \textit{Modality Translation} evaluates its capability to bridge severe cross-domain gaps while maintaining strict anatomical consistency; and (3) \textit{Super Resolution} assesses its capability to recover high-frequency, fine-grained details from severe structural information loss. 
As shown in Table \ref{tab:translation_all}, DDB consistently outperforms baselines, achieving significantly higher PSNR and SSIM, alongside lower LPIPS scores across all three settings. These results confirm that the explicit spatial anchors injected via our hybrid absorption process successfully empower the model to mitigate severe domain shifts and execute highly accurate structural mapping. As shown in Fig. \ref{fig:i2i_translation}, DDB successfully restores complex textures from degraded images, and generates highly realistic target modalities with sharp anatomical boundaries in CT-to-MRI translation, proving the robustness and precision of our spatiotemporally aligned diffusion trajectory. More results of the I2I task are in Appendix.

\subsection{Performance on T2I Generation Tasks}
\label{sec:T2I task}
\begin{figure*}[t]
    \centering
        \centering
        \includegraphics[width=0.8\textwidth]{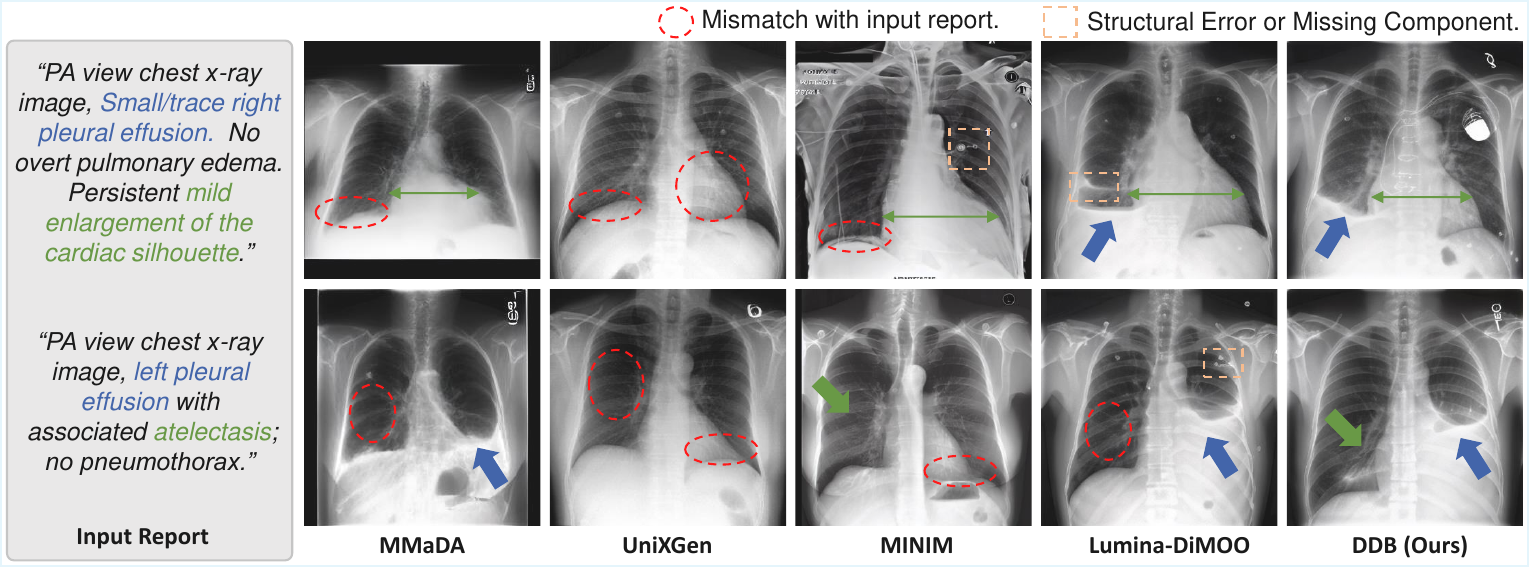}
        \caption{Visual comparison of different methods and DDB for report-to-image generation task on the MIMIC-CXR dataset.}
        \label{fig:t2i_report_image}
\end{figure*}

\input{table/t2i}
To investigate the generation capability of our model without source images as explicit spatial source anchors, we perform a report-to-image generation (T2I) task.
This task requires generating coherent X-ray images strictly from complex medical text descriptions. As shown in Table \ref{tab:t2i}, DDB achieves superior FID, MS-SSIM, and CLIP-Score compared to other methods. Visual comparisons in Fig. \ref{fig:t2i_report_image} further show that our generated X-rays exhibit more realistic anatomical features and fewer structural artifacts. This confirms that introducing spatiotemporal alignment does not compromise, but rather enhances, the model's fundamental generative priors.

\subsection{Component Analysis of DDB Framework}
\label{sec:analysis}
To better understand the core design principles of DDB and build an effective model, we analyze how different components affect the effectiveness of our discrete diffusion process. We focus on the following key questions:
\begin{itemize}
\item Source injection ratio: How does the hybrid injection ratio balance structural preservation against editing alignment during forward process? (Table \ref{tab:injection_ratio})
\item Source injection mechanism: Which spatial assignment strategy (stochastic mixing or deterministic allocation) yields the most robust hybrid absorbing state? (Table \ref{tab:inject_where})
\item Information metrics: How do varying definitions of information impact scheduling effectiveness across different tasks? (Table \ref{tab:information})
\end{itemize}
\subsubsection{Source injection ratio.}
We analyze the impact of the source token injection ratio during the forward process. As shown in Table \ref{tab:injection_ratio}, an injection ratio of 0.5 combined with random-ratio training yields superior performance. We draw three key conclusions. First, a larger injection ratio yields better source and background preservation (higher DINO), whereas a smaller ratio favors stronger edit semantic alignment (higher CLIP-T). Second, an injection ratio of 0.5, where noise and source tokens are evenly mixed, achieves the optimal balance between CLIP-T and DINO, resulting in the highest overall Edit Score. This validates the efficacy of hybrid noising. Furthermore, training with a random ratio significantly outperforms any fixed ratio configuration. This occurs because random ratio training exposes the model to diverse contextual combinations, thereby enhancing its robustness and generalization capabilities during inference. This leads to a key insight:
\textbf{Takeaway 1.} \textit{An optimal balance of source and mask tokens (ratio around 0.5) coupled with random-ratio training maximizes both structural preservation and editing flexibility.}

\input{table/source_injection_ratio}

\subsubsection{Source injection mechanism.}
\input{table/inject_where}
We evaluate three injection mechanisms for corrupted regions to determine the best strategy for mixing source and mask tokens: (1) Deterministically injecting mask tokens to high-information regions and source tokens to low-information regions; (2) The inverse of (1); (3) Randomly assigning source or mask tokens based on a fixed ratio. Empirical results in Table \ref{tab:inject_where} demonstrate that the stochastic proportional mix (M3) achieves the best performance. This happens because deterministic allocation often causes the model to learn fixed mappings, excessively limiting the flexibility of generation. In contrast, stochastic injection forces the network to robustly learn transition mappings under diverse and unpredictable contextual guidance. This leads to a key insight:
\textbf{Takeaway 2.} \textit{Stochastic proportional injection is superior because it ensures contextual robustness and effectively decouples the stochastic spatial state from the deterministic temporal schedule.}

\vspace{-4mm}
\subsubsection{Information metrics.}
We study the influence of different information metrics used to guide our masking schedule. As shown in Table \ref{tab:information}, for translation tasks utilizing explicit source images (e.g., Image Editing, Style Transfer), Image Difference achieves highest performance. Conversely, for unanchored generation tasks lacking source images (e.g., Report-to-Image), Target variance emerges as the optimal metric. This indicates that structural translation relies on relative cross-domain changes, whereas pure generation depends on image complexity to quantify generation difficulty. This leads to a key insight:
\textbf{Takeaway 3.} \textit{The optimal information metric is task-dependent: latent difference is crucial for anchored image-to-image translation, while target variance is optimal for unanchored text-to-image generation.}
\input{table/information_map}

\subsection{Performance on Few-Step Inference Setting}
\label{sec:few-step}
\input{table/few_step}
We analyze the model's performance retention across varying sampling steps to evaluate its inference efficiency and robustness. Standard discrete diffusion models typically suffer severe performance degradation when inference steps are heavily reduced, primarily due to compounded prediction errors and the lack of explicit guidance. As shown in Table \ref{tab:few_step}, while the baseline model exhibits a significant performance drop at low step counts (5 steps), DDB maintains robust generation quality and editing accuracy. This demonstrates that our spatiotemporally aligned trajectory effectively simplifies the learning objective, enabling highly efficient and accurate decoding even under accelerated inference settings.

\section{Conclusion}
We propose Discrete Diffusion Bridges (DDB), a novel framework that resolves the fundamental spatiotemporal misalignment inherent in standard discrete diffusion models. Our work reconceptualizes the discrete generative trajectory from a random, void-to-image process into a direct, efficient bridge between data domains. By introducing hybrid absorption to establish robust spatial source anchors and an information-guided schedule to dictate an optimal, difficulty-aware temporal path, DDB perfectly synchronizes training objectives with inference dynamics. Extensive experiments demonstrate DDB's versatility and superiority across a wide range of tasks. Our work opens new perspectives on optimal trajectory design in discrete latent spaces and provides novel insights into modeling unified multimodal translation and generation.

\section*{Acknowledgments}
This work was supported by the National Natural Science Foundation of China (U23A20343), the National Natural Science Foundation of China under Grant (T2594604030, T2594604035, 62306253), the Early Career Fund (27207025), the National Natural Science Foundation of China under Grant (U24A20282), the Guangdong Natural Science Fund-General Program (2024A1515010233), the China Postdoctoral Science Foundation under Grant Number 2025M781669, and the Fundamental Research Project of SIA (2025JC1K05).

\bibliographystyle{splncs04}
\bibliography{main}

\clearpage
\appendix

\setcounter{section}{0}
\renewcommand{\thesection}{S\arabic{section}}
\renewcommand{\thesubsection}{\thesection.\arabic{subsection}}
\renewcommand{\thesubsubsection}{\thesubsection.\arabic{subsubsection}}

\begin{center}
    {\large\bfseries Discrete Diffusion Bridges for Spatiotemporally Aligned\\
    Image Translation and Generation}
    
    \vspace{2mm}
    
    {\large\bfseries Supplementary Material}
\end{center}

\vspace{6mm}

\input{Supplementary_Material_arxiv.tex}

\end{document}

%% file: table/edit_omni_bench.tex
\begin{table*}[!t] 
  \small
  \centering
  \renewcommand{\arraystretch}{1.25}
  \setlength{\tabcolsep}{1.5mm}  
  \caption{Performance comparisons of Image Editing and Style Transfer tasks on the OmniEdit \cite{wei2024omniedit} benchmark, and Sub-driven generation tasks on the Graph-200K \cite{li2025visualcloze} benchmark. Best results are highlighted in \textcolor{red}{\textbf{red}} and the second-best results are \textcolor{blue}{blue}.}
  \resizebox{1.0\linewidth}{!}{
  \begin{tabular}{l ccc cc ccc}
    \toprule
     \multirow{2}{*}{\textbf{Method}}  &  \multicolumn{3}{c}{\textbf{Image Editing}}  & \multicolumn{2}{c}{\textbf{Style Transfer}} & \multicolumn{3}{c}{\textbf{Sub-Driven Generation}} \\
    \cmidrule(lr){2-4} \cmidrule(lr){5-6} \cmidrule(lr){7-9}& Edit Score $\uparrow$ & CLIP-T$\uparrow$& DINO $\uparrow$ & Edit Score $\uparrow$ & CLIP-T $\uparrow$& CLIP-I$\uparrow$& CLIP-T $\uparrow$& DINO $\uparrow$ \\ \hline
     
     MMaDA \cite{yang2025mmada}& -0.233& 0.231& 0.643& -0.188& 0.244& -& -& -\\
     
     OmniGen \cite{xiao2025omnigen}& 0.129& 0.242& 0.636& 0.587& 0.246& 0.828& 0.343& 0.710\\
     
     AnyEdit \cite{yu2024anyedit}& 0.110& 0.239& 0.762& 0.319& 0.228& 0.736& 0.319& 0.542\\
 Bagel \cite{deng2025emerging}& 0.694& 0.233& \textcolor{red}{\textbf{0.836}}& 0.128& 0.242& \textcolor{blue}{\textbf{0.858}}& 0.348&0.745\\
     
     OmniGen2 \cite{wu2025omnigen2}& \textcolor{blue}{\textbf{0.752}}& \textcolor{blue}{\textbf{0.240}}& 0.784& 0.628& 0.240& 0.843& 0.344& 0.739\\

     Lavida-O \cite{li2025lavida}& 0.619& \textcolor{red}{\textbf{0.242}}& 0.681& \textcolor{red}{\textbf{1.233}}& 0.246& 0.856& \textcolor{red}{\textbf{0.359}}& \textcolor{blue}{\textbf{0.774}}\\

     Lumina. \cite{xin2025lumina}& 0.734& 0.234& 0.766& 0.958& \textcolor{blue}{\textbf{0.246}}& 0.843& 0.344& 0.767\\
     
     \textbf{DDB (Ours)}& \textcolor{red}{\textbf{0.779}}& 0.236& \textcolor{blue}{\textbf{0.790}}& \textcolor{blue}{\textbf{1.077}}& \textcolor{red}{\textbf{0.250}}& \textcolor{red}{\textbf{0.867}}& \textcolor{blue}{\textbf{0.349}}& \textcolor{red}{\textbf{0.789}}\\
    \bottomrule
  \end{tabular}
  }
  \label{tab:style_subject_editing}
\end{table*}

%% file: table/I2I_translation_all.tex
\begin{table*}[!t] 
  \small
  \centering
  \renewcommand{\arraystretch}{1.25}
  \setlength{\tabcolsep}{1.5mm}  
  \caption{Evaluate All-in-One Restoration, Modality translation, and Super Resolution tasks on CDD-11 \cite{guo2024onerestore}, SynthRAD \cite{thummerer2023synthrad2023}, and IXI \cite{ixi_dataset} datasets, respectively.}
  \resizebox{1.0\linewidth}{!}{
  \begin{tabular}{l ccc ccc ccc}
    \toprule
     \multirow{2}{*}{\textbf{Method}}  &\multicolumn{3}{c}{\textbf{All-in-One Restoration}}  &  \multicolumn{3}{c}{\textbf{Modality Translation}
}& \multicolumn{3}{c}{\textbf{Super Resolution}
} \\
    \cmidrule(lr){2-4} \cmidrule(lr){5-7} \cmidrule(lr){8-10} 
&PSNR $\uparrow$&SSIM $\uparrow$& LPIPS $\downarrow$&  PSNR $\uparrow$&SSIM $\uparrow$&LPIPS $\downarrow$
& PSNR $\uparrow$&SSIM $\uparrow$&LPIPS $\downarrow$
\\ \hline
     
     MmaDA \cite{yang2025mmada}&17.92& 0.595& 0.187&  22.24&0.742& 0.110
& 22.89& 0.704& 0.087
\\
 Muddit\cite{shi2025muddit}&20.29& 0.608& 0.140& 21.50& 0.585& 0.164& 23.75& 0.648&0.109\\
 Lumina. \cite{xin2025lumina}&21.96& 0.751& 0.092& 23.05& 0.732& 0.095
& 25.15& 0.719&0.066
\\
     
     \textbf{DDB (Ours)}&\textcolor{red}{\textbf{22.61}}& \textcolor{red}{\textbf{0.759}}& \textcolor{red}{\textbf{0.081}}&  \textcolor{red}{\textbf{23.36}}&\textcolor{red}{\textbf{0.750}}& \textcolor{red}{\textbf{0.083}}& \textcolor{red}{\textbf{25.46}}& \textcolor{red}{\textbf{0.733}}& \textcolor{red}{\textbf{0.056}}\\ \hline
  \end{tabular}
     }
  \label{tab:translation_all}
  \vspace{-2mm}
\end{table*}

%% file: table/t2i.tex
\begin{wraptable}{r}{6.3cm}
  \vspace{-10mm}
  \small
  \centering
  \renewcommand{\arraystretch}{1.25}
  \setlength{\tabcolsep}{1.5mm}  
  \caption{Evaluate Report to Image generation task on MIMIC-CXR \cite{johnson2019mimic} datasets.}
  \resizebox{1.0\linewidth}{!}{
  \begin{tabular}{l ccc}
    \toprule
     \multirow{2}{*}{\textbf{Method}}  & \multicolumn{3}{c}{\textbf{Report-to-Image}} \\
    \cmidrule(lr){2-4} 
& FID $\downarrow$&MS-SSIM $\uparrow$&CLIP-Score $\uparrow$\\ \hline
     
     MmaDA \cite{yang2025mmada}& 15.25& 0.381& 0.446\\
     
     UniXGen \cite{lee2023vision}& 30.75& 0.361& 0.413\\
 MINIM \cite{wang2025self}& 15.62& 0.317&0.442\\
 Lum.-DiMOO \cite{xin2025lumina}& 13.86& 0.398&0.441\\
     
     \textbf{DDB (Ours)}& \textcolor{red}{\textbf{9.81}}& \textcolor{red}{\textbf{0.413}}& \textcolor{red}{\textbf{0.451}}\\ \hline
  \end{tabular}
     }
  \label{tab:t2i}
  \vspace{-8mm}
\end{wraptable}

%% file: table/source_injection_ratio.tex
\begin{table}[t]
  \small
  \centering
  \renewcommand{\arraystretch}{1.25}
  \setlength{\tabcolsep}{1.5mm} 
  \caption{Impact of source token injection ratios on the OmniEdit benchmark, comparing models trained with a fixed ratio against trained with a random ratio ($r \in [0, 1]$).}
  \resizebox{0.75\linewidth}{!}{
  \begin{tabular}{l cccccc}
    \toprule
     \multirow{2}{*}{\textbf{Ratio}}&  \multicolumn{3}{c}{\textbf{Train on specified ratio}} & \multicolumn{3}{c}{\textbf{Train on random ratio}}\\
    \cmidrule(lr){2-4} \cmidrule(lr){5-7}  & Edit Score $\uparrow$ & CLIP-T$\uparrow$& DINO $\uparrow$  & Edit Score $\uparrow$ & CLIP-T$\uparrow$&DINO $\uparrow$  \\ \hline
     
     0.0& 0.743& 0.235& 0.755& 0.760& 0.235&0.765\\
     
     0.2& 0.745& 0.233& 0.761& 0.770& 0.234&0.774\\
     
     0.5& 0.757& 0.235& 0.778& \textbf{0.779}& \textbf{0.236}&\textbf{0.790}\\
 0.8& 0.756& 0.234& 0.774& 0.768& 0.233&0.785\\
     
     1.0& 0.750& 0.233& 0.781& 0.764& 0.232&0.788\\ \hline
  \end{tabular}
  }
  \label{tab:injection_ratio}
  \vspace{-1mm}
\end{table}

%% file: table/inject_where.tex
\begin{table*}[t] 
  \small
  \centering
  \renewcommand{\arraystretch}{1.25}
  \setlength{\tabcolsep}{1.5mm}  
  \caption{The impact of varying source injection mechanisms. We compare deterministic allocation (M1: masking high-information regions; M2: the inverse) against stochastic proportional mixing (M3).}
  \resizebox{1.0\linewidth}{!}{
  \begin{tabular}{l ccc cc lccc}
    \toprule
     \multirow{2}{*}{\textbf{Metric}}  &  \multicolumn{3}{c}{\textbf{Image Editing}}  & \multicolumn{3}{c}{\textbf{Modality Translation}}& \multicolumn{3}{c}{\textbf{Super Resolution}} \\
    \cmidrule(lr){2-4} \cmidrule(lr){5-7} \cmidrule(lr){8-10}& Edit Score $\uparrow$ & CLIP-T$\uparrow$& DINO $\uparrow$ & PSNR $\uparrow$& SSIM $\uparrow$&LPIPS $\downarrow$& PSNR $\uparrow$& SSIM $\uparrow$& LPIPS $\downarrow$\\ \hline
 w/o inj.& 0.743& 0.235& 0.755& 23.15& 0.741&0.089& 25.19& 0.723&0.059\\

     M1& 0.695& 0.231& 0.783& 21.96& 0.714&0.102& 23.59& 0.688& 0.073\\

     M2& 0.747& 0.234& 0.781& 23.11& 0.746&0.088& 25.10& 0.725& 0.061\\
 M3& \textbf{0.779}& \textbf{0.236}& \textbf{0.790}& \textbf{23.36}& \textbf{0.750}&\textbf{0.083}& \textbf{25.46}& \textbf{0.733}&\textbf{0.056}\\ \hline
  \end{tabular}
  }
  \label{tab:inject_where}
  \vspace{-5mm}
\end{table*}

%% file: table/information_map.tex
\begin{table*}[!t] 
  \small
  \centering
  \renewcommand{\arraystretch}{1.25}
  \setlength{\tabcolsep}{1.5mm}  
  \caption{The impact of using different information metrics on image editing, style transfer, and report-to-image generation tasks.}
  \resizebox{1.0\linewidth}{!}{
  \begin{tabular}{l ccc cc ccc}
    \toprule
     \multirow{2}{*}{\textbf{Metric}}  &  \multicolumn{3}{c}{\textbf{Image Editing}}  & \multicolumn{2}{c}{\textbf{Style Transfer}} & \multicolumn{3}{c}{\textbf{Report-to-Image}} \\
    \cmidrule(lr){2-4} \cmidrule(lr){5-6} \cmidrule(lr){7-9}& Edit Score $\uparrow$ & CLIP-T$\uparrow$& DINO $\uparrow$ & Edit Score $\uparrow$ & CLIP-T$\uparrow$& FID $\downarrow$& MS-SSIM $\uparrow$& CLIP-Score $\uparrow$\\ \hline
 w/o infor.& 0.755& 0.235& 0.778& 0.982& 0.245& 14.88& 0.397&0.441\\

     Gradient& 0.686& 0.236& 0.764& 0.979& 0.250& 11.50& 0.403& 0.450\\

     Frequency& 0.758& 0.235& 0.780& 1.069& 0.247& 11.59& 0.410& 0.449\\
 Variance& 0.723& 0.236& 0.772& 0.983& 0.249& \textbf{9.81}& \textbf{0.413}&\textbf{0.451}\\
     
     Image Diff& \textbf{0.779}& \textbf{0.236}& \textbf{0.790}& \textbf{1.077}& \textbf{0.250}& -& -& -\\
    \bottomrule
  \end{tabular}
  }
  \label{tab:information}
  \vspace{-1mm}
\end{table*}

%% file: table/few_step.tex
\begin{table}[t]
  \small
  \centering
  \renewcommand{\arraystretch}{1.25}
  \setlength{\tabcolsep}{1.5mm}  
  \caption{Quantitative evaluation under few-step inference settings.Our DDB maintains robust performance even at extremely low sampling steps.}
  \resizebox{0.8\linewidth}{!}{
  \begin{tabular}{l ccc ccc}
    \toprule
     \multirow{2}{*}{\textbf{Step}}&  \multicolumn{3}{c}{\textbf{Lumina-DiMOO}} & \multicolumn{3}{c}{\textbf{DDB(Ours)}}\\
    \cmidrule(lr){2-7}& Edit Score $\uparrow$ & CLIP-Acc $\uparrow$& DINO $\uparrow$  & Edit Score $\uparrow$ & CLIP-Acc $\uparrow$&DINO $\uparrow$  \\ \hline
 5 steps& 0.541& 0.233& 0.743& 0.606& 0.235&0.757\\ 
     
     15 steps& 0.709& 0.235&  0.756& 0.735& 0.235&0.762\\

     30 steps& 0.716& 0.234&  0.758
& 0.763& 0.235&0.769\\

     45 steps& 0.721& 0.235&  0.757& 0.776& 0.235&0.770\\
     
     64 steps& 0.734& 0.234& 0.766& \textbf{0.779}& \textbf{0.236}&\textbf{0.790}\\
    \bottomrule
  \end{tabular}
  }
  \label{tab:few_step}
  \vspace{-2mm}
\end{table}

%% file: Supplementary_Material_arxiv.tex
\section{Overview}
Our supplementary materials include the following sections:
\begin{itemize}
    \item More implementation details, including datasets and experiment settings. (Section~\ref{sec: implementation}).
    \item Additional experimental results, including more results on I2I translation, further analysis of DDB framework, and more ablation study. (Section~\ref{sec: experimental});
    \item More visual results. (Section~\ref{sec: visual});
\end{itemize}

\section{More Implementation Details}
\label{sec: implementation}
\subsection{Datasets}

\subsubsection{Image Editing and Style Transfer.}
We adopt the OmniEdit \cite{wei2024omniedit} dataset to evaluate instruction-based image editing and style transfer. It provides paired source-target images with precise text instructions covering diverse tasks, including object manipulation (swap, removal, addition), attribute modification, background/environment changes, and global style transfer. We utilize 1.2M image pairs for training and evaluate on the benchmark comprising 700 test cases.

\subsubsection{Subject-Driven Generation.}
We utilize the Graph-200K \cite{li2025visualcloze} dataset for the subject-driven generation task. It provides paired isolated foreground subjects and complex background contexts, assessing the model's ability to seamlessly contextualize background-free targets into new text-described environments. Our split includes 182,000 image pairs for training and 1,000 pairs for testing.

\subsubsection{All-in-One Restoration.} 
For the All-in-One restoration task, we employ the CDD-11 \cite{guo2024onerestore} dataset. It features 11 composite degradation combinations (e.g., low light and haze; haze and rain; low light, haze, and snow) to evaluate restoration under complex, mixed real-world scenarios. We use the standard split of 20,790 training pairs and 2,310 testing pairs.

\subsubsection{Modality Translation.}
To evaluate pure structural mapping across severe domain gaps without textual guidance, we use the SynthRAD2023 \cite{thummerer2023synthrad2023} dataset for CT-to-MRI and MRI-to-CT translation. It contains strictly aligned, paired CT and MRI scans from identical patients. We use 83,864 image pairs for training and 555 pairs for testing. We followed the pipeline introduced in HealthGPT \cite{lin2025healthgpt} for processing.

\subsubsection{Super Resolution.}
We employ the IXI dataset \cite{ixi_dataset} of normal brain MR images for the super-resolution task, assessing the recovery of fine-grained structural details from degraded, low-resolution inputs. We use 44,551 image pairs for training and 600 pairs for testing. We followed the pipeline introduced in HealthGPT \cite{lin2025healthgpt} for processing.

\subsubsection{Report-to-Image.}
We use the MIMIC-CXR \cite{johnson2019mimic} dataset, which contains a large number of real X-ray images and medical report pairs, to evaluate medical report-to-image generation. We extract three radiographic views: posteroanterior (PA), anteroposterior (AP), and lateral (LATENT). The text prompts are automatically formatted using the diagnostic report's impression section as: ``\{view\} view chest X-ray image, \{impression\}''. We select 221,238 pairs for training and 1,000 pairs for testing.

\subsubsection{Image Inpainting.}
We use the CelebA-HQ \cite{karras2017progressive} (CelebFaces High Quality) dataset for the image inpainting task. CelebA-HQ consists of 30,000 high-resolution facial images. The dataset includes a wide variety of celebrity faces with various attributes such as age, gender, and facial expressions, making it suitable for tasks like face generation and image inpainting. Each image in the dataset is paired with a mask that specifies the region to be inpainted, allowing models to learn to fill in missing parts of the face. Following IR-SDE \cite{luo2023image}, we select 29901 pairs for training and 99 pairs for testing.

\subsection{Experiment Settings}
For DDB models, all experiments are implemented using PyTorch on 4 NVIDIA A100 GPUs. We employ a batch size of 8 per GPU with a learning rate of 2e-5, optimized by AdamW \cite{loshchilov2017decoupled} (weight decay=0.1, $\beta_{1}$=0.9, $\beta_{2}$=0.95). We use Lumina DiMOO \cite{xin2025lumina} as the backbone of the DDB framework, following its model settings and conducting full-scale parameter training. All images are preprocessed through center cropping to 512×512 pixels. All data are pretokenized before training to increase throughput. The VQ tokenizer operates with a downsampling rate of 16, resulting in a 1024-token representation for each image.
For other comparison methods, we use their official implementation for training and inference.


\section{Additional Experimental Results}
\label{sec: experimental}
\subsection{More Results on I2I Translation}
\subsubsection{Editing Performance across Diverse Sub-Tasks.} 
To provide a more granular evaluation of our DDB framework, we compare its performance against established baselines across various image editing sub-tasks within the OmniEdit \cite{wei2024omniedit} benchmark. As detailed in Table \ref{tab:sub_task}, our method consistently achieves the highest overall Edit Score across nearly all evaluated sub-tasks. This comprehensive superiority further validates the efficacy of DDB in optimally balancing precise semantic edit alignment with strict source context preservation, regardless of the specific editing instruction.
\input{table/edit_sub_task}

\subsubsection{Mask Inpainting.} 
To further validate the generalization capabilities of our proposed framework across a broader spectrum of I2I tasks, we evaluate its performance on image inpainting using the CelebA-HQ \cite{karras2017progressive} dataset. Specifically, we utilize irregular masks, which more accurately simulate the complex and unpredictable occlusions frequently encountered in real-world scenarios. As reported in Table \ref{tab:image_inpainting}, DDB consistently outperforms baseline methods. This underscores the robustness of our spatiotemporally aligned trajectory in recovering coherent semantic and structural information even under severe spatial corruption.
\input{table/image_inpainting}


\subsubsection{Unified Image Translation and Generation.} 
To validate the versatility of our framework, we evaluate its performance under a unified training paradigm, where a single model with shared parameters simultaneously handles diverse image translation and generation tasks. Specifically, we jointly train the model on a comprehensive suite of benchmarks: OmniEdit \cite{wei2024omniedit} for image editing and style transfer, Graph-200K \cite{li2025visualcloze} for subject-driven generation, CDD-11 \cite{guo2024onerestore} for all-in-one restoration, SynthRAD2023 \cite{thummerer2023synthrad2023} for modality translation, IXI \cite{ixi_dataset} for super-resolution, MIMIC-CXR \cite{johnson2019mimic} for report-to-image synthesis, and CelebA-HQ \cite{karras2017progressive} for image inpainting. As shown in Table \ref{tab:unified}, DDB consistently outperforms the baseline across all tasks under this unified setting. This demonstrates that our framework effectively accommodates multi-task learning, establishing a robust foundation for universal visual generation.

\input{table/unified}

\subsubsection{Compare with related bridge works.}
We compare DDB with relevant bridge-style I2I methods in Table~\ref{tab:comparison_rw}, further demonstrating the advantage of DDB.

\begin{table}[t]
  \small
  \centering
  \renewcommand{\arraystretch}{1.25}
  \setlength{\tabcolsep}{1.5mm}  
  \caption{Performance comparison with bridge-style methods for I2I translation.}
  \resizebox{\textwidth}{!}{
  \begin{tabular}{c ccc}
    \toprule
     \multirow{2}{*}{\textbf{Method}}&\textbf{All-in-One Restoration}&  \textbf{Modality translation}& \textbf{Super Resolution}\\
     \cmidrule(lr){2-4}
    &PSNR/SSIM/LPIPS&  PSNR/SSIM/LPIPS& PSNR/SSIM/LPIPS\\ \hline
     
     DDIB \cite{su2022dual}&15.76/0.532/0.273&  14.59/0.443/0.312& 15.25/0.498/0.297\\
 BBDM \cite{li2023bbdm}&20.21/0.549/0.126& 19.69/0.546/0.229& 21.82/0.640/0.105\\
 DDBM \cite{zhou2024denoising}&22.38/0.704/0.092& 21.39/0.609/0.171& 24.34/0.721/0.061\\
     
     \textbf{DDB (Ours)}&\textbf{22.61/0.759/0.081}&  \textbf{23.36/0.750/0.083}& \textbf{25.46/0.733/0.056}\\ \hline
  \end{tabular}
     }
  \label{tab:comparison_rw}
\end{table}

\begin{figure*}[h]
    \centering
    \includegraphics[width=\textwidth]{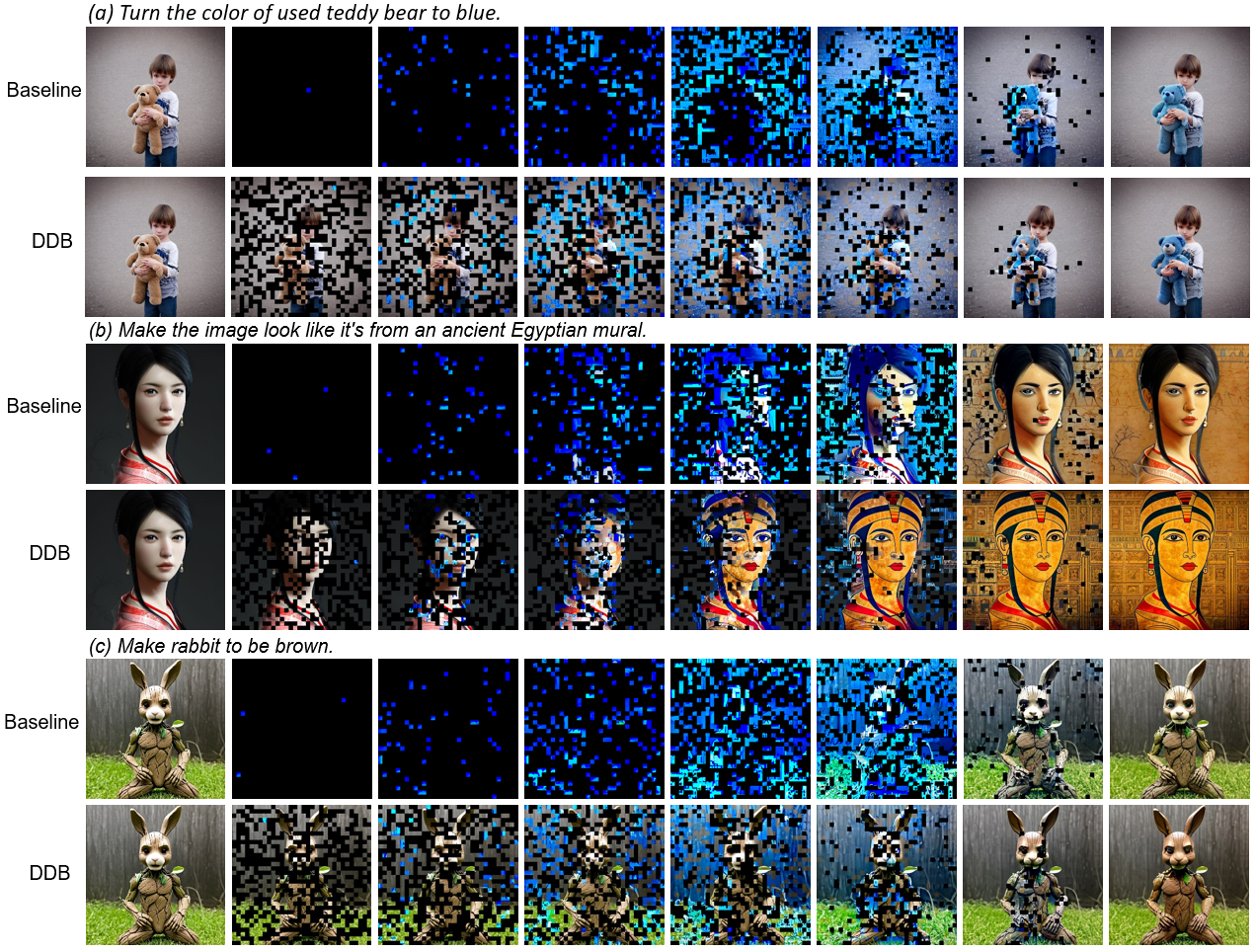}
    \caption{Compare the inference process using the baseline and our DDB framework on the OmniEdit \cite{wei2024omniedit} dataset. Using Lumin-DiMOO \cite{xin2025lumina} as the baseline model. Our method can more effectively translate input images to the target domain. (a)(c) image editing, (b) style transfer.}
    \label{fig:inference_main}
\end{figure*}
\subsection{Further Analysis of DDB Framework}

\begin{figure*}[h]
    \centering
    \includegraphics[width=\textwidth]{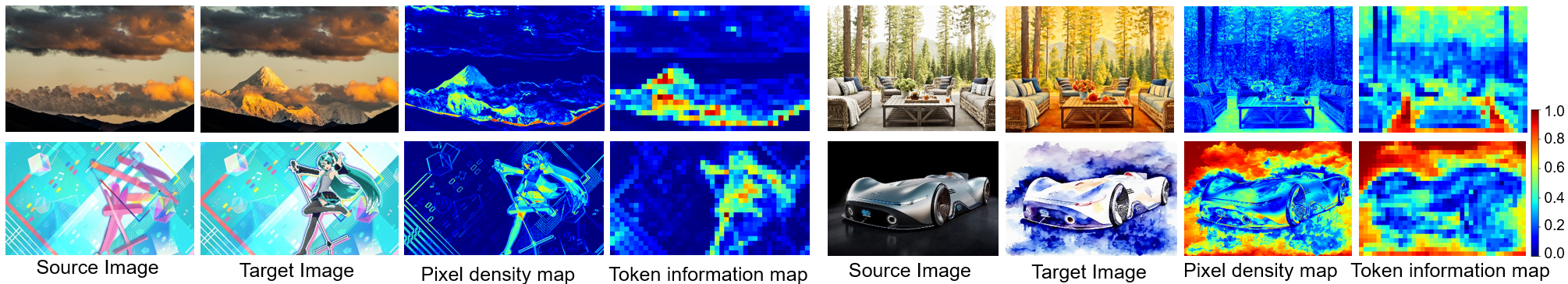}
    \caption{Calculate the pixel-level density map and token-level information map between the source image and the target image, using edit difference as the information metric.}
    \label{fig:infor_supp}
\end{figure*}

\subsubsection{Analysis of Alignment Mechanism between Training and Inference.}
To gain deeper insights into the discrete diffusion process, we explore its intrinsic decoding dynamics. As shown in Fig. \ref{fig:inference_main}, empirical observations reveal a pronounced \textbf{"easy-first, hard-last" decoding trajectory}. For regional manipulation tasks, such as the image editing example in Fig. \ref{fig:inference_main}(a), the model prioritizes decoding the invariant background before synthesizing the edited subject (the "teddy bear"). Conversely, in global style transfer tasks (Fig. \ref{fig:inference_main}(b)), the model first reconstructs the source-consistent primary subject (the person), leaving the complex, style-variant background to be decoded later. Both phenomena indicate that \textbf{discrete diffusion models inherently resolve simple, high-confidence contexts before tackling difficult, semantically altered regions.}

Motivated by this observation, we designed the \textbf{information-guided masking schedule} to explicitly align the training objective with this inference behavior. As shown in Fig. \ref{fig:infor_supp}, our mechanism accurately quantifies regions with substantial cross-domain variations as "high-information" areas. By prioritizing the corruption of these regions during the forward process, we match the "hard-to-decode" areas identified during inference. Simultaneously, we \textbf{leverage the source image as an explicit spatial anchor}, significantly narrowing the spatial distance between the source and target domains. As further illustrated in Fig. \ref{fig:inference_main}, where the blue regions represent perturbation tokens that typically emerge when insufficient tokens are decoded at early steps, our method drastically mitigates this decoding interference. Driven by this spatial anchor mechanism, the model acquires robust source spatial priors at an early stage, revealing the spatial state of the target domain much faster and enabling a highly stable cross-domain transition.

This synergistic spatiotemporal alignment empowers the model with a superior ability to localize and execute modifications. As depicted in Fig. \ref{fig:inference_main}(c), the standard baseline model struggles to capture the correct spatial regions for semantic manipulation, often leading to translation failures. In contrast, our aligned framework precisely localizes the intended subjects for modification, ensuring accurate and robust generation in the target domain.



\subsection{More Ablation Study}
\subsubsection{Stochastic Masking Ratio.}
\input{table/stochastic}
We studied the effect of stochastic masking ratio $\rho$, set different ratios from 0 to 0.5 for model training, and compared the performance. As shown in Table \ref{tab:stochastic}, the optimal value is reached when the stochastic masking ratio $\rho$=0.3.


\section{More Visual Comparisons}
\label{sec: visual}

\begin{figure*}[h]
    \centering
    \includegraphics[width=\textwidth]{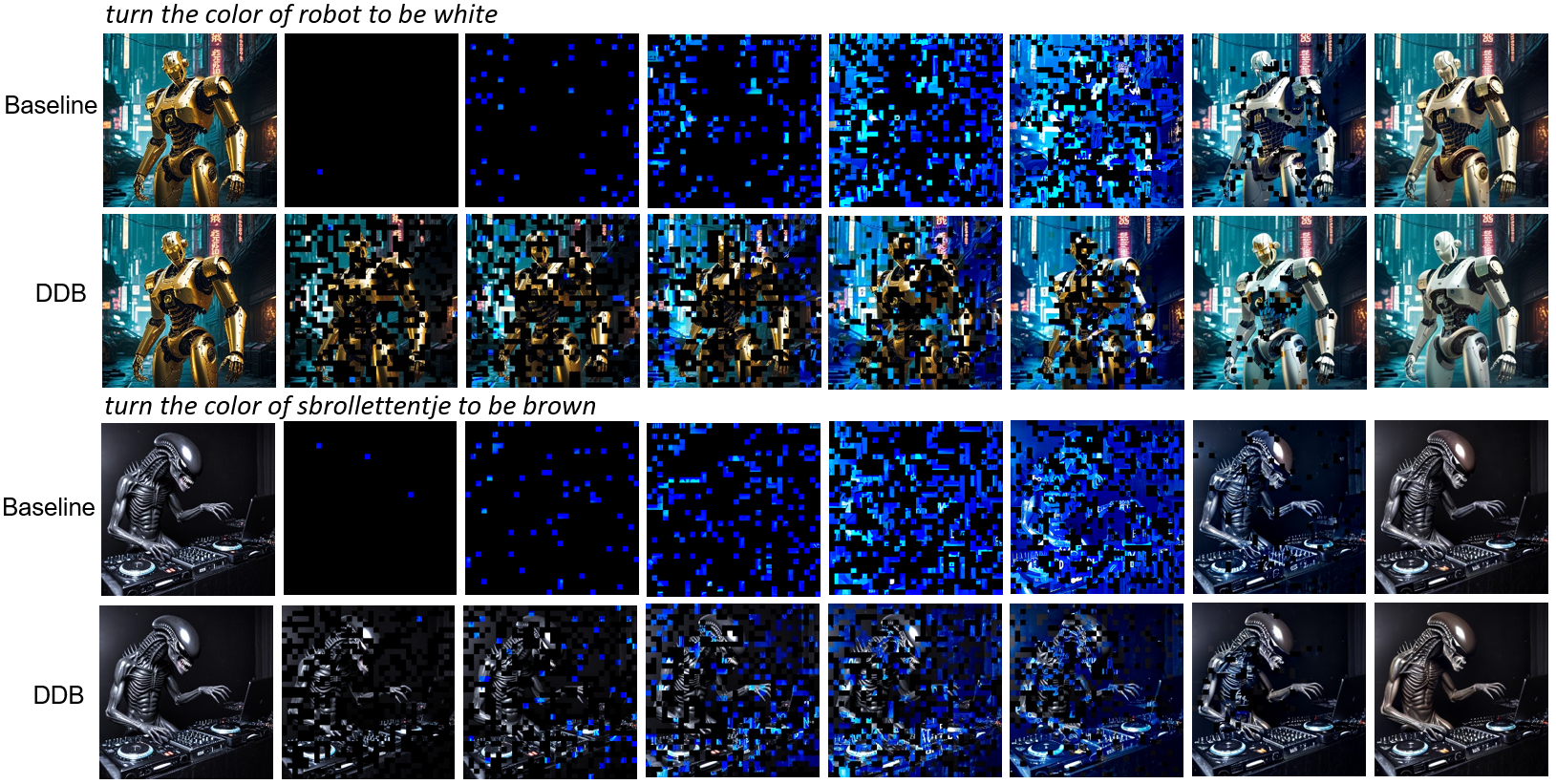}
    \caption{Visual comparison of the baseline and our DDB framework in the \textit{attribution modification} task on the OmniEdit \cite{wei2024omniedit} dataset. Use Lumin-DiMOO \cite{xin2025lumina} as the baseline model. Our method can more effectively translate input images to the target domain.}
    \label{fig:inference_color}
\end{figure*}

\begin{figure*}[h]
    \centering
    \includegraphics[width=\textwidth]{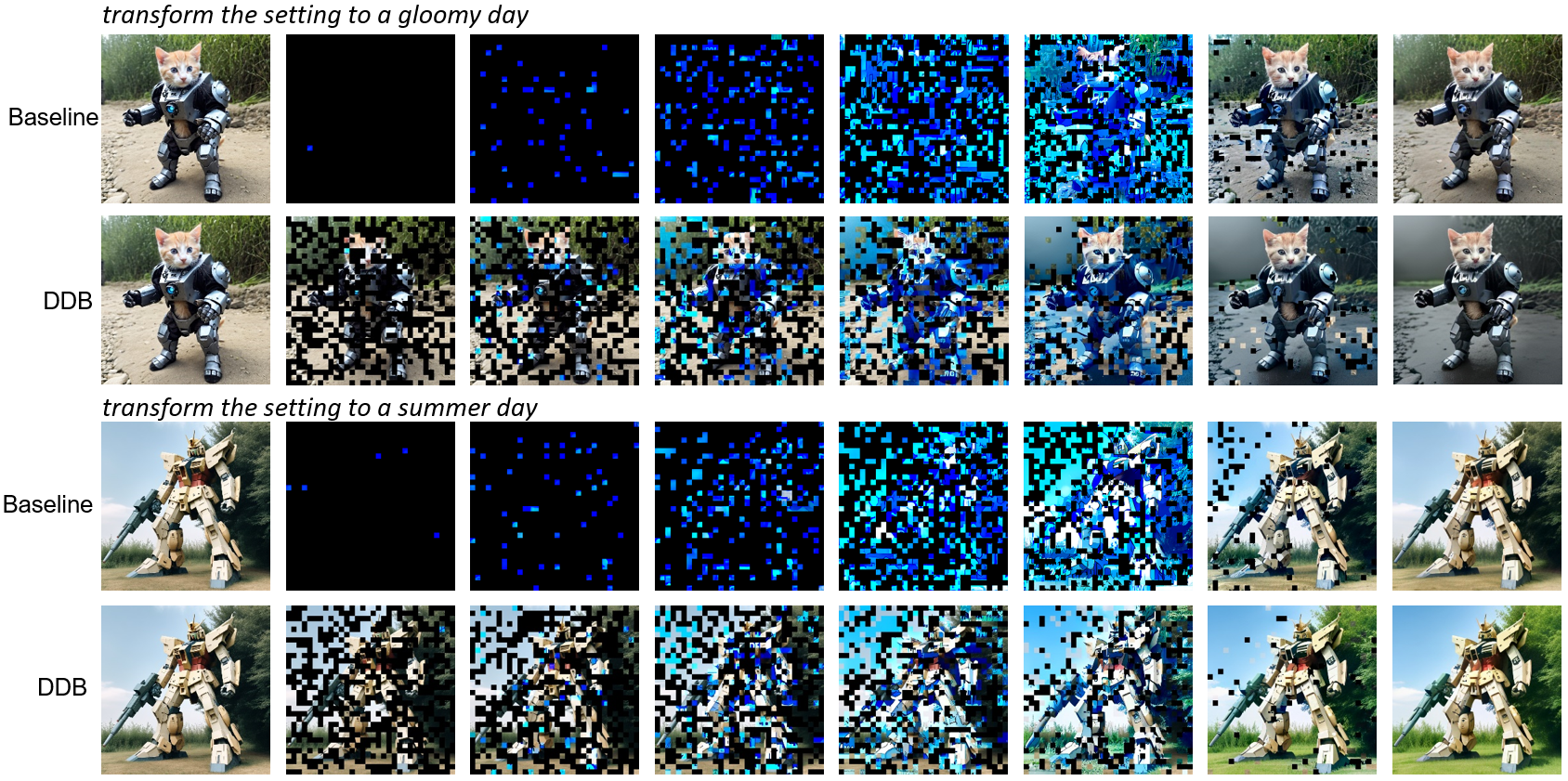}
    \caption{Visual comparison of the baseline and our DDB framework in the \textit{weather change} task on the OmniEdit \cite{wei2024omniedit} dataset. Use Lumin-DiMOO \cite{xin2025lumina} as the baseline model. Our method can more effectively translate input images to the target domain.}
    \label{fig:inference_weather}
\end{figure*}

\begin{figure*}[h]
    \centering
    \includegraphics[width=\textwidth]{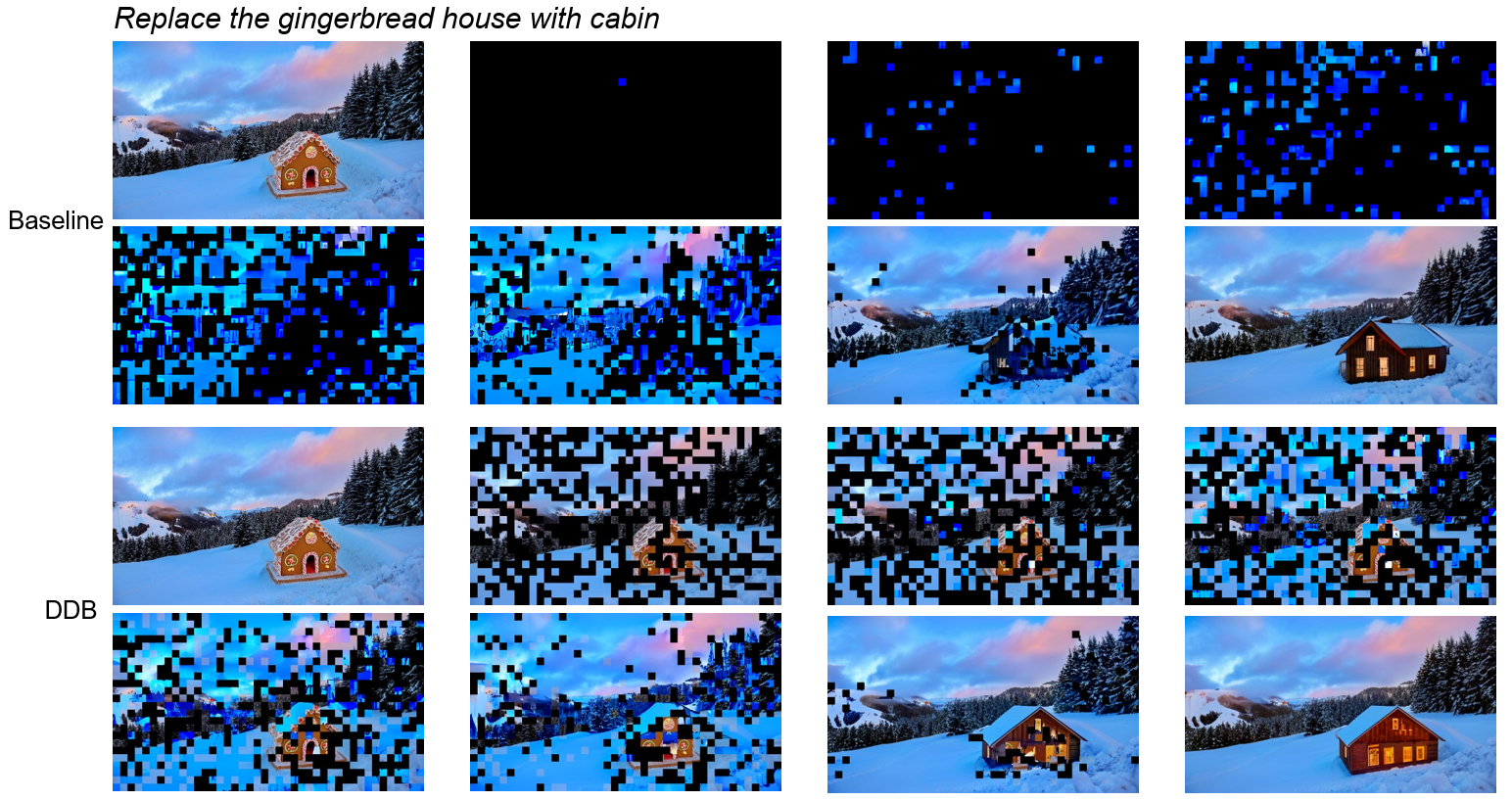}
    \caption{Visual comparison of the baseline and our DDB framework in the \textit{object swap} task on the OmniEdit \cite{wei2024omniedit} dataset. Use Lumin-DiMOO \cite{xin2025lumina} as the baseline model. Our method can more effectively translate input images to the target domain.}
    \label{fig:inference_replace1}
\end{figure*}

\begin{figure*}[h]
    \centering
    \includegraphics[width=\textwidth]{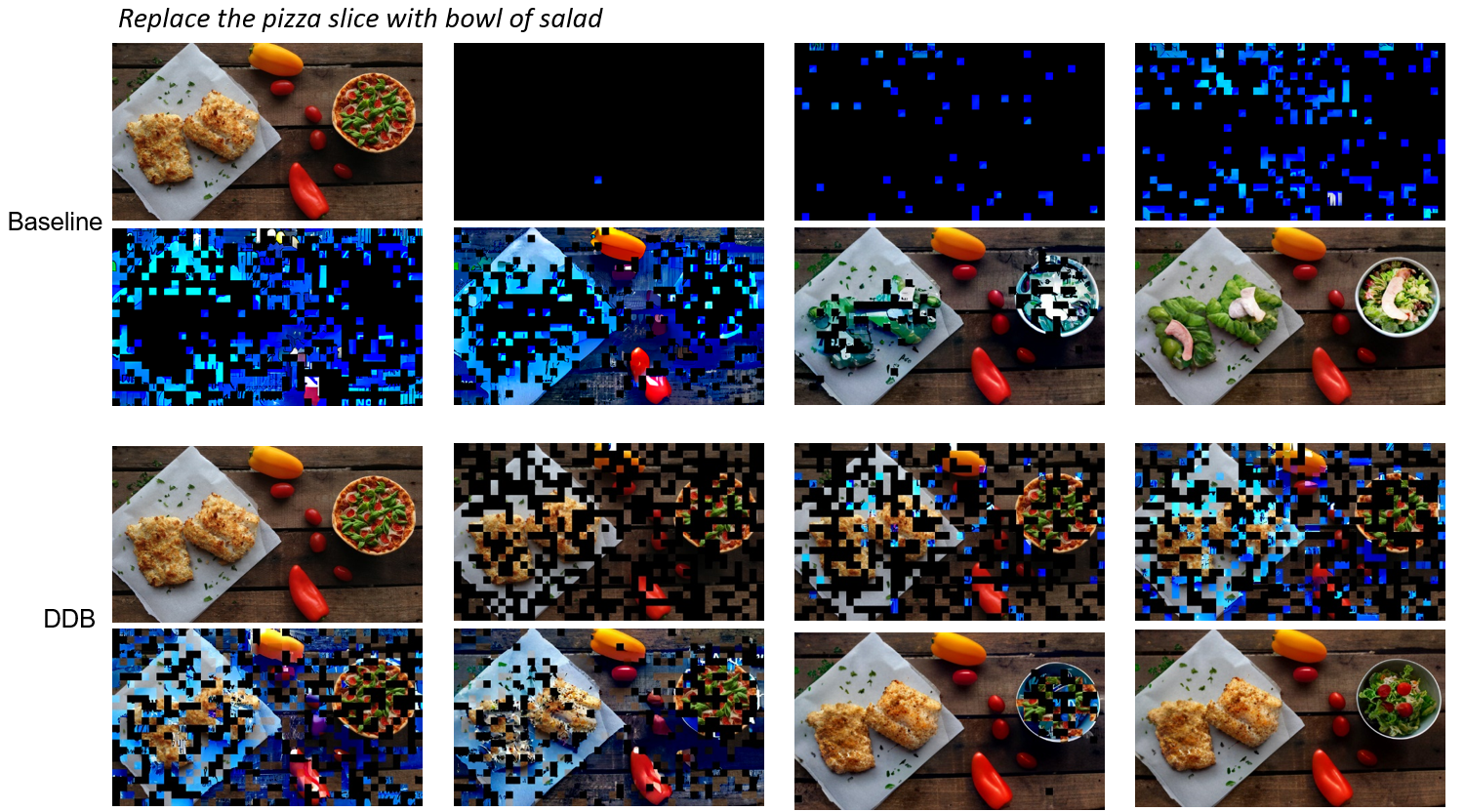}
    \caption{Visual comparison of the baseline and our DDB framework in the \textit{object swap} task on the OmniEdit \cite{wei2024omniedit} dataset. Use Lumin-DiMOO \cite{xin2025lumina} as the baseline model. Our method can more effectively translate input images to the target domain.}
    \label{fig:inference_replace2}
\end{figure*}

\begin{figure*}[h]
    \centering
    \includegraphics[width=\textwidth]{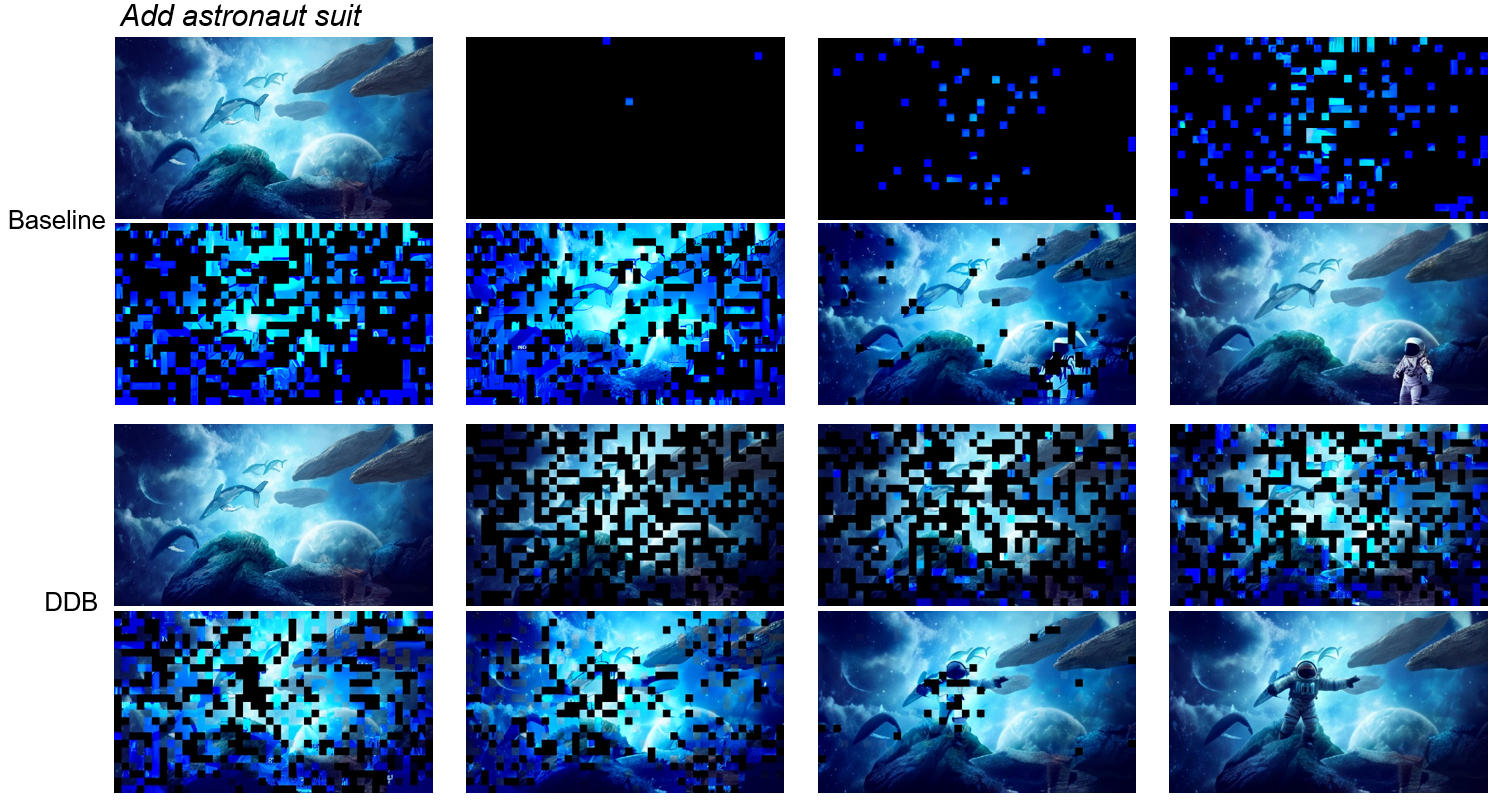}
    \caption{Visual comparison of the baseline and our DDB framework in the \textit{object add} task on the OmniEdit \cite{wei2024omniedit} dataset. Use Lumin-DiMOO \cite{xin2025lumina} as the baseline model. Our method can more effectively translate input images to the target domain.}
    \label{fig:inference_add}
\end{figure*}

\begin{figure*}[h]
    \centering
    \includegraphics[width=\textwidth]{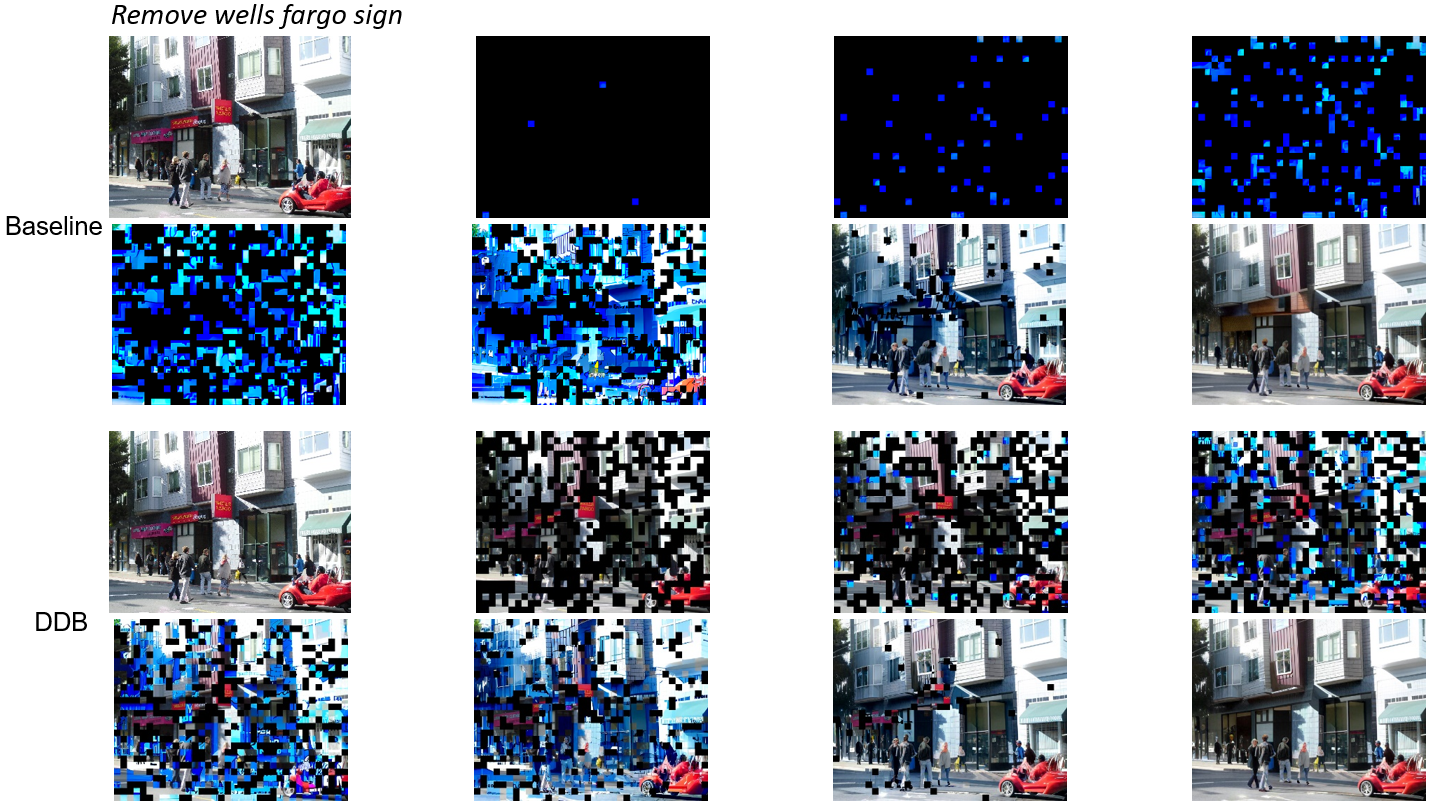}
    \caption{Visual comparison of the baseline and our DDB framework in the \textit{object remove} task on the OmniEdit \cite{wei2024omniedit} dataset. Use Lumin-DiMOO \cite{xin2025lumina} as the baseline model. Our method can more effectively translate input images to the target domain.}
    \label{fig:inference_remove}
\end{figure*}

\begin{figure*}[h]
    \centering
    \includegraphics[width=\textwidth]{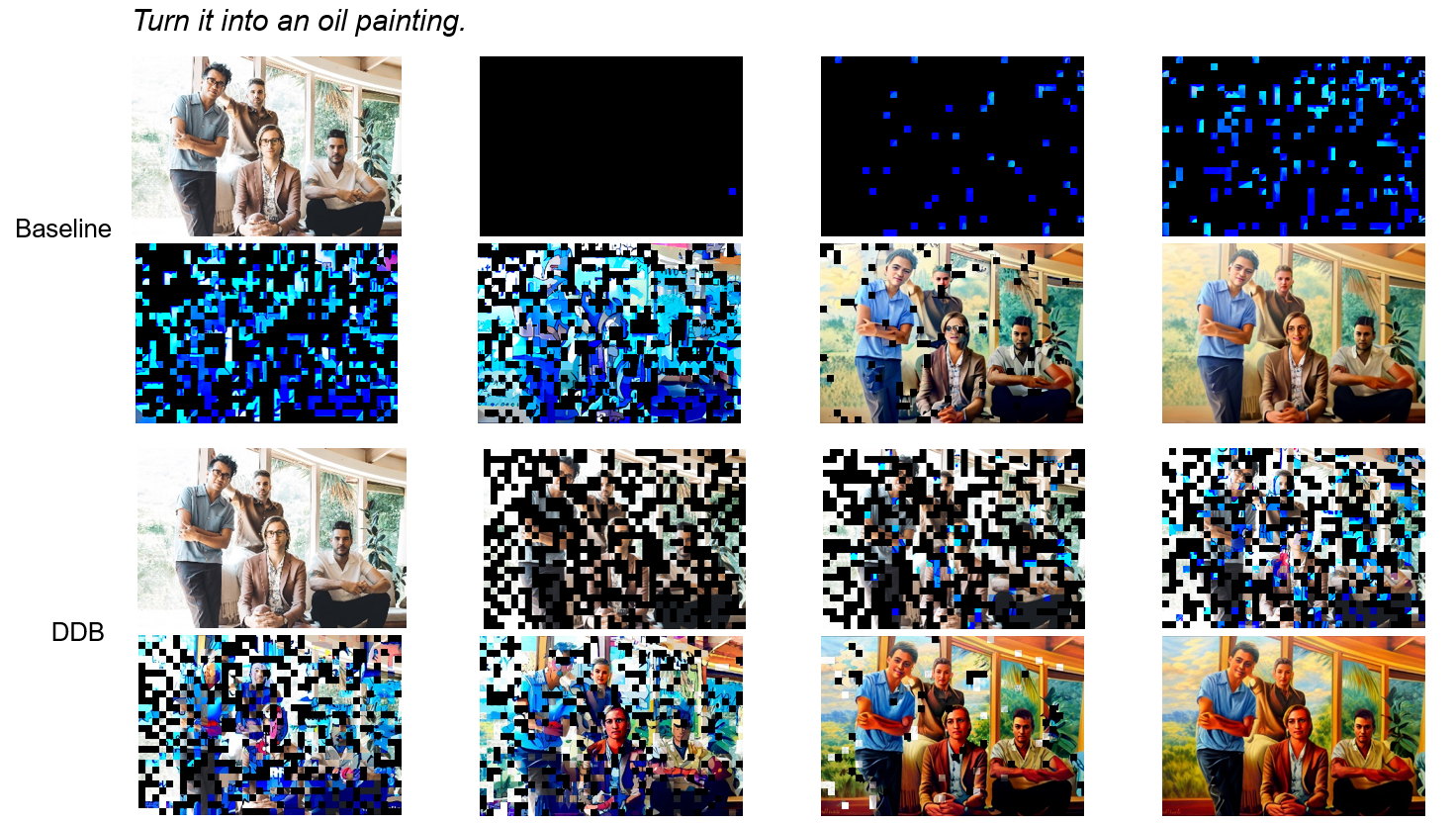}
    \caption{Visual comparison of the baseline and our DDB framework in the \textit{style transfer} task on the OmniEdit \cite{wei2024omniedit} dataset. Use Lumin-DiMOO \cite{xin2025lumina} as the baseline model. Our method can more effectively translate input images to the target domain.}
    \label{fig:inference_style}
\end{figure*}

\begin{figure*}[h]
    \centering
    \includegraphics[width=\textwidth]{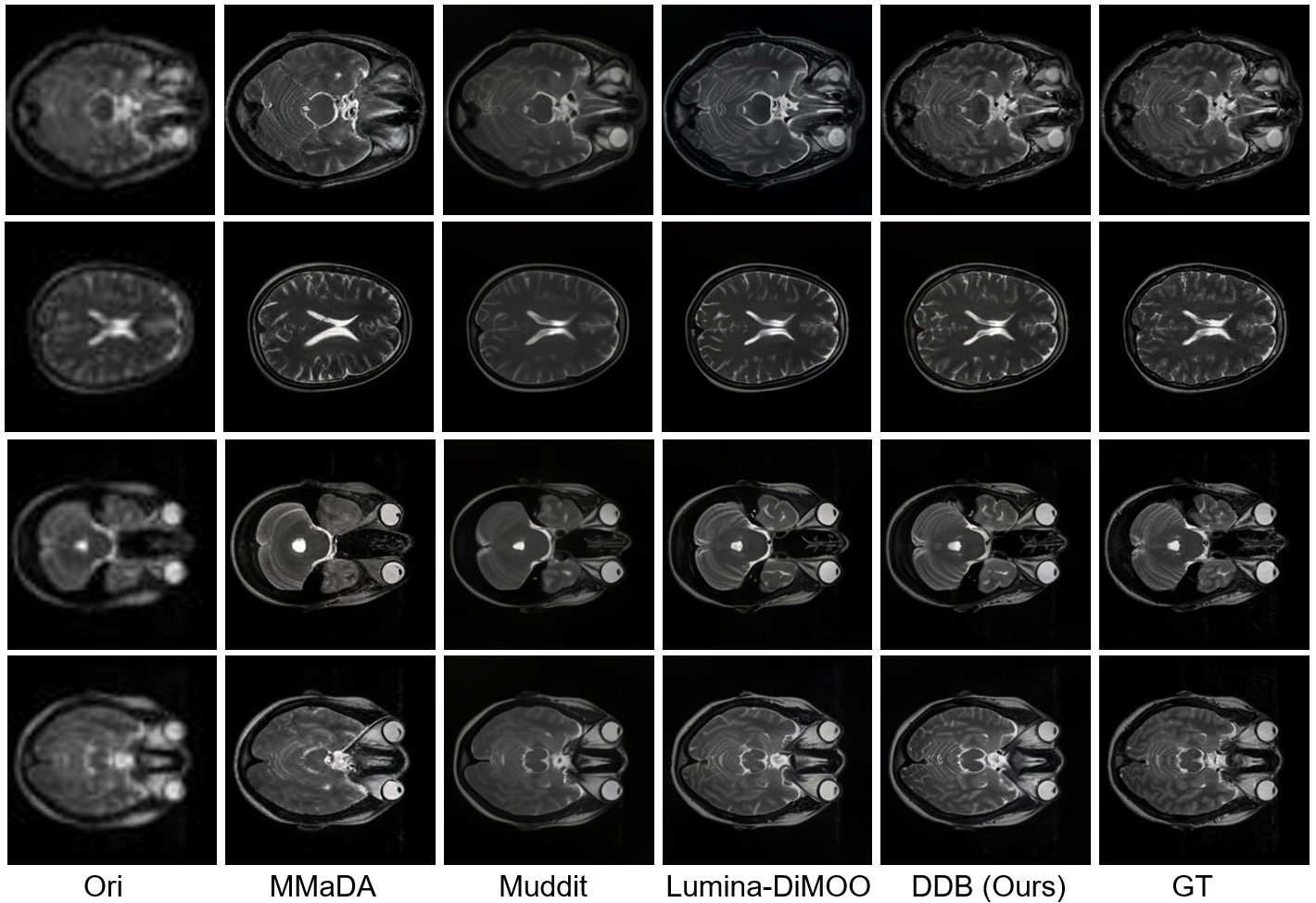}
    \caption{Visual comparison in the super resolution task on the IXI \cite{ixi_dataset} dataset.}
    \label{fig:sr_supp}
\end{figure*}

\begin{figure*}[h]
    \centering
    \includegraphics[width=\textwidth]{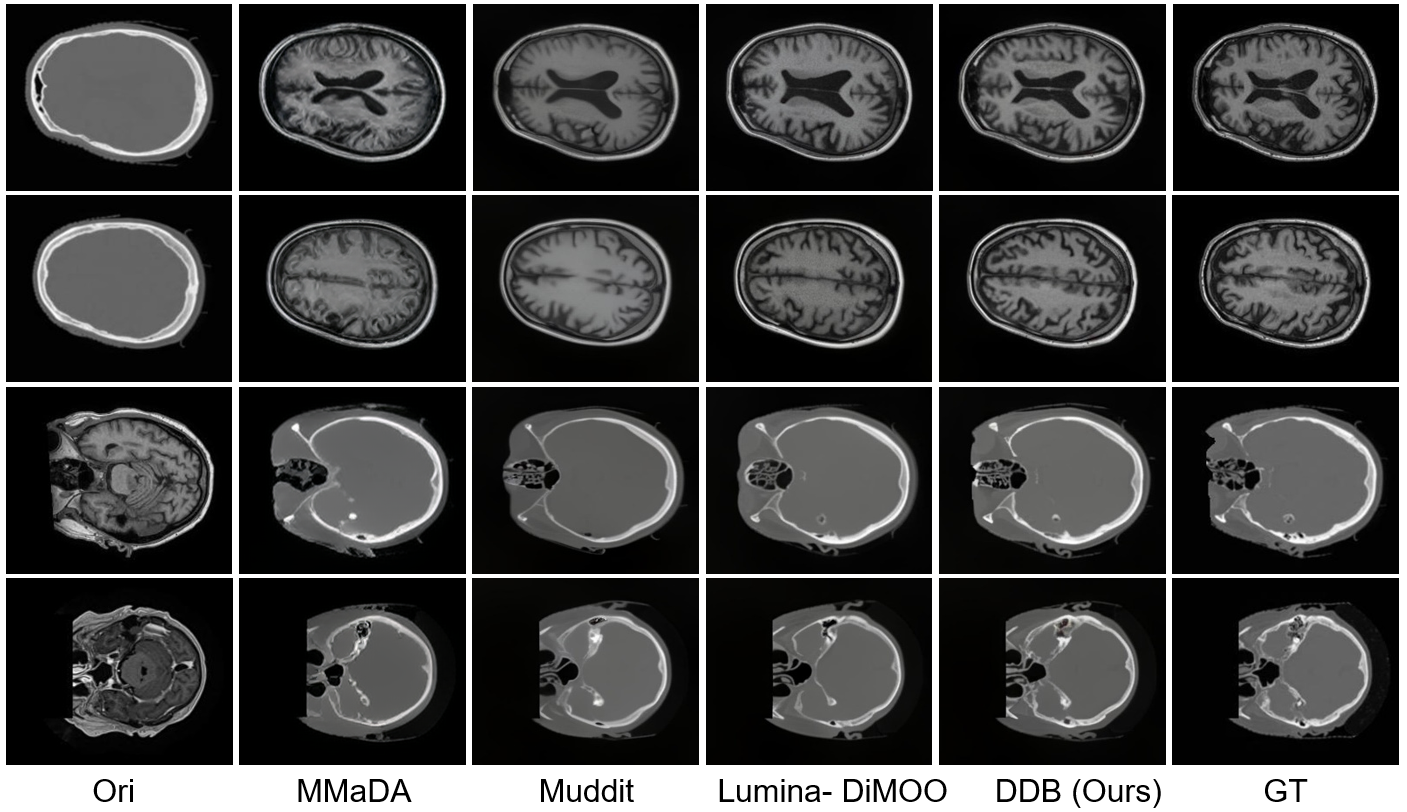}
    \caption{Visual comparison in the modality translation task on the SynthRAD2023 \cite{thummerer2023synthrad2023} dataset.}
    \label{fig:MT_Supp}
\end{figure*}

\begin{figure*}[h]
    \centering
    \includegraphics[width=\textwidth]{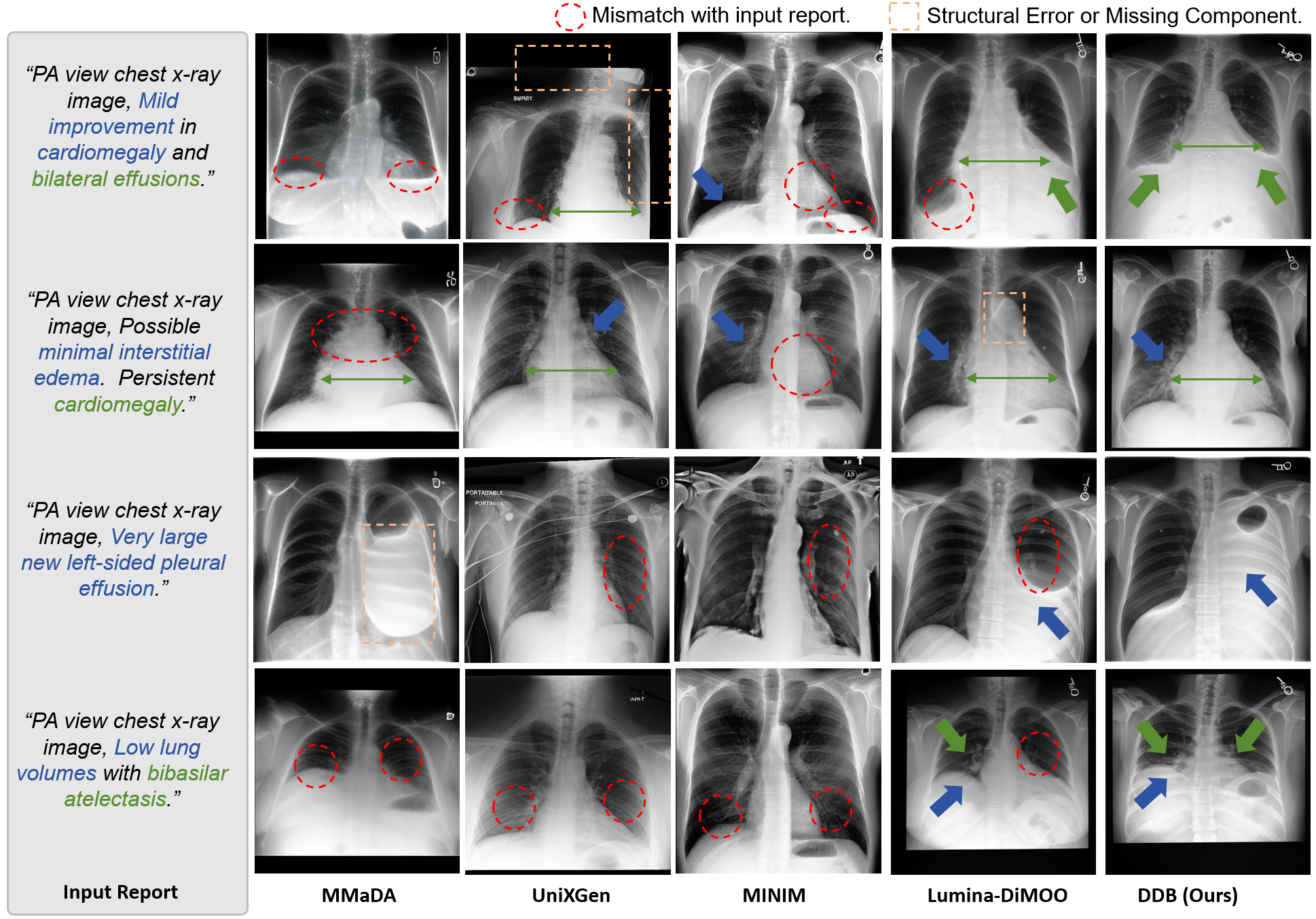}
    \caption{Visual comparison in the report-to-image generation task on the MIMIC-CXR \cite{johnson2019mimic} dataset.}
    \label{fig:xray_supp}
\end{figure*}

We present additional visual results from experiments, including the following:
\begin{itemize}
    \item Visual comparisons of the baseline and our DDB framework on the OmniEdit \cite{wei2024omniedit} dataset. Attribution modification: Fig. \ref{fig:inference_color}, weather change: Fig. \ref{fig:inference_weather}, object swap: Fig. \ref{fig:inference_replace1} and Fig. \ref{fig:inference_replace2}, object add: \ref{fig:inference_add}, object remove: \ref{fig:inference_remove}, style transfer \ref{fig:inference_style};
    \item Visual comparison in the super resolution task. (Fig.~\ref{fig:sr_supp});
    \item Visual comparison in the modality translation task. (Fig.~\ref{fig:MT_Supp});
    \item Visual comparison in the report-to-image generation task. (Fig.~\ref{fig:xray_supp}).
\end{itemize}

%% file: table/edit_sub_task.tex
\begin{table*}[!t] 
  \small
  \centering
  \renewcommand{\arraystretch}{1.25}
  \setlength{\tabcolsep}{1.5mm}  
  \caption{Performance comparison of various image editing subtasks on the OmniEdit \cite{wei2024omniedit} benchmark. Best results are highlighted in \textcolor{red}{\textbf{red}} and the second-best results are \textcolor{blue}{\textbf{blue}}. We achieved the \textbf{best overall performance} (highest Edit Score) on almost all subtasks.}
  \resizebox{1.0\linewidth}{!}{
  \begin{tabular}{l ccc ccc ccc}
    \toprule
     \multirow{2}{*}{\textbf{Method}}  &  \multicolumn{3}{c}{\textbf{Object Swap}}  & \multicolumn{3}{c}{\textbf{Object Removal}}& \multicolumn{3}{c}{\textbf{Object Addition}} \\
    \cmidrule(lr){2-4} \cmidrule(lr){5-7} \cmidrule(lr){8-10}
    & Edit Score $\uparrow$ & CLIP-T $\uparrow$& DINO $\uparrow$ & Edit Score $\uparrow$ & CLIP-T $\uparrow$&DINO $\uparrow$ & Edit Score $\uparrow$ & CLIP-T $\uparrow$& DINO $\uparrow$ \\ \midrule
     
     MMaDA \cite{yang2025mmada}& -0.048& 0.231& 0.584& -0.386& 0.209&0.581& -0.875& 0.227& 0.596\\
     OmniGen \cite{xiao2025omnigen}& 0.123& 0.242& 0.573& -0.601& 0.210& 0.531& -0.126& 0.227&0.699\\
     AnyEdit \cite{yu2024anyedit}& -0.316& 0.247& 0.677& -0.272& 0.210& 0.716& -0.142& 0.223&0.767\\
     Bagel \cite{deng2025emerging}& 0.967& 0.240& 0.701& \textcolor{blue}{\textbf{0.456}}& 0.212& 0.772& \textcolor{blue}{\textbf{0.256}}& 0.214&\textcolor{red}{\textbf{0.912}}\\
     OmniGen2 \cite{wu2025omnigen2}& \textcolor{red}{\textbf{1.181}}& \textcolor{red}{\textbf{0.303}}& 0.705& 0.439& \textcolor{red}{\textbf{0.217}}& 0.758& 0.223& 0.224&0.781\\
     Lavida-O \cite{li2025lavida}& 0.653& 0.239& 0.648& 0.208& 0.210& 0.705& 0.147& \textcolor{red}{\textbf{0.235}}&0.637\\
     Lumina. \cite{xin2025lumina}& 0.935& 0.262& \textcolor{blue}{\textbf{0.709}}& 0.264& 0.210& \textcolor{blue}{\textbf{0.778}}& 0.209& 0.227&0.780\\
     \rowcolor{gray!20}
     \textbf{DDB (Ours)}& \textcolor{blue}{\textbf{0.995}}& \textcolor{blue}{\textbf{0.270}}& \textcolor{red}{\textbf{0.713}}& \textcolor{red}{\textbf{0.523}}& \textcolor{blue}{\textbf{0.211}}&\textcolor{red}{\textbf{0.792}}& \textcolor{red}{\textbf{0.307}}& \textcolor{blue}{\textbf{0.228}}& \textcolor{blue}{\textbf{0.794}}\\
     
    \midrule
     \multirow{2}{*}{\textbf{Method}}  &  \multicolumn{3}{c}{\textbf{Attribute Modification}}  & \multicolumn{3}{c}{\textbf{Background Swap}}& \multicolumn{3}{c}{\textbf{Environment Change}} \\
    \cmidrule(lr){2-4} \cmidrule(lr){5-7} \cmidrule(lr){8-10}
    & Edit Score $\uparrow$ & CLIP-T $\uparrow$& DINO $\uparrow$ & Edit Score $\uparrow$ & CLIP-T $\uparrow$&DINO $\uparrow$ & Edit Score $\uparrow$ & CLIP-T $\uparrow$& DINO $\uparrow$ \\ \midrule
    
     MMaDA \cite{yang2025mmada}& -0.233& 0.280& 0.753& -0.174& 0.224&0.562& 0.178& 0.227& 0.702\\
     OmniGen \cite{xiao2025omnigen}& 0.183& 0.281& 0.757& 0.294& \textcolor{blue}{\textbf{0.239}}& 0.575& 0.886& 0.228&0.681\\
     AnyEdit \cite{yu2024anyedit}& 0.239& \textcolor{red}{\textbf{0.297}}& 0.817& 0.164& \textcolor{red}{\textbf{0.241}}& 0.610& 0.815& 0.215&0.823\\
     Bagel \cite{deng2025emerging}& 0.502& 0.282& 0.858& 1.047& 0.236& \textcolor{red}{\textbf{0.791}}& 0.709& 0.217&\textcolor{red}{\textbf{0.883}}\\
     OmniGen2 \cite{wu2025omnigen2}& 0.612& 0.288& 0.823& 1.007& 0.225& \textcolor{blue}{\textbf{0.732}}& 1.023& 0.227&0.826\\
     Lavida-O \cite{li2025lavida}& \textcolor{blue}{\textbf{0.638}}& \textcolor{blue}{\textbf{0.289}}& 0.791& 0.942& 0.235& 0.630& \textcolor{blue}{\textbf{1.220}}& \textcolor{red}{\textbf{0.244}}&0.746\\
     Lumina. \cite{xin2025lumina}& 0.604& 0.282& \textcolor{blue}{\textbf{0.867}}& \textcolor{blue}{\textbf{1.075}}& 0.216& 0.675& 1.113& 0.222&0.829\\
     \rowcolor{gray!20}
     \textbf{DDB (Ours)}& \textcolor{red}{\textbf{0.641}}& 0.282& \textcolor{red}{\textbf{0.869}}& \textcolor{red}{\textbf{1.098}}& 0.227& 0.689& \textcolor{red}{\textbf{1.252}}& \textcolor{blue}{\textbf{0.231}}& \textcolor{blue}{\textbf{0.832}}\\
    \bottomrule
  \end{tabular}
  }
  \label{tab:sub_task}
\end{table*}

%% file: table/image_inpainting.tex
\begin{table*}[!t] 
  \small
  \centering
  \renewcommand{\arraystretch}{1.25}
  \setlength{\tabcolsep}{1.5mm}  
  \caption{Evaluate the image inpainting task on CelebA-HQ \cite{karras2017progressive} dataset.}
  \resizebox{0.5\linewidth}{!}{
  \begin{tabular}{l ccc }
    \toprule
     \multirow{2}{*}{\textbf{Method}}  &\multicolumn{3}{c}{\textbf{Image Inpainting}}  \\
    \cmidrule(lr){2-4} 
&PSNR $\uparrow$&SSIM $\uparrow$& LPIPS $\downarrow$\\ \hline
     
     MmaDA \cite{yang2025mmada}&21.90& 0.753& 0.142\\
 Muddit\cite{shi2025muddit}&24.74& 0.776& 0.108\\
 Lumina. \cite{xin2025lumina}&25.85& 0.806& 0.081\\
     \rowcolor{gray!20}
     \textbf{DDB (Ours)}&\textbf{26.26}& \textbf{0.811}& \textbf{0.069}\\ \toprule
  \end{tabular}
     }
  \label{tab:image_inpainting}
  \vspace{-2mm}
\end{table*}

%% file: table/unified.tex
\begin{table*}[!t] 
  \small
  \centering
  \renewcommand{\arraystretch}{1.25}
  \setlength{\tabcolsep}{1.5mm}  
  \caption{Performance comparison during unified training for all tasks.}
  \resizebox{1.0\linewidth}{!}{
  \begin{tabular}{l ccc ccc cccccc}
    \toprule
     \multirow{2}{*}{\textbf{Method}}  &  \multicolumn{3}{c}{\textbf{Image Editing}}  & \multicolumn{3}{c}{\textbf{Style Transfer}}& \multicolumn{3}{c}{\textbf{Sub-Driven Generation}} &  \multicolumn{3}{c}{\textbf{All-In-One}}\\
    \cmidrule(lr){2-4} \cmidrule(lr){5-7} \cmidrule(lr){8-10} \cmidrule(lr){11-13}
    & Edit Score&CLIP-T &DINO& \multicolumn{2}{c}{Edit Score}&CLIP-T
& CLIP-I &CLIP-T  &DINO & PSNR &SSIM  &LPIPS \\ \hline
     Lumina. \cite{xin2025lumina}& 0.727& 0.234& \textbf{0.788}& \multicolumn{2}{c}{0.924}&
0.246& 0.865& 0.350& 0.773& 20.93& 0.751&0.116\\
     \rowcolor{gray!20}
     \textbf{DDB (Ours)}& \textbf{0.758}& \textbf{0.235}& 0.771& \multicolumn{2}{c}{\textbf{1.106}}&\textbf{0.247}& \textbf{0.865}& \textbf{0.351}&  \textbf{0.776}& \textbf{21.69}& \textbf{0.752}&\textbf{0.095}\\
     
    \midrule
     \multirow{2}{*}{\textbf{Method}}  &  \multicolumn{3}{c}{\textbf{Modality Translation}}  & \multicolumn{3}{c}{\textbf{Super Resolution}}& \multicolumn{3}{c}{\textbf{Report-to-Image}} &  \multicolumn{3}{c}{\textbf{Image Inpainting}}\\
    \cmidrule(lr){2-4} \cmidrule(lr){5-7} \cmidrule(lr){8-10} \cmidrule(lr){11-13}
    & PSNR &SSIM  &LPIPS& PSNR &SSIM &LPIPS & FID &MS-SSIM &CLIP-T& PSNR &SSIM  &LPIPS\\\hline
     Lumina. \cite{xin2025lumina}& 21.40& 0.716& 0.127& 24.47& 0.728& 0.078& 15.77& 0.403& 0.437& 25.21& 0.781&0.085\\
     \rowcolor{gray!20}
     \textbf{DDB (Ours)}& \textbf{21.93}& \textbf{0.723}& \textbf{0.105}& \textbf{24.91}& \textbf{0.736}& \textbf{0.064}& \textbf{11.14}& \textbf{0.423}&  \textbf{0.443}& \textbf{25.78}& \textbf{0.785}&\textbf{0.076}\\
    \bottomrule
  \end{tabular}
  }
  \label{tab:unified}
\end{table*}

%% file: table/stochastic.tex
\begin{table*}[!t] 
  \small
  \centering
  \renewcommand{\arraystretch}{1.25}
  \setlength{\tabcolsep}{1.5mm}  
  \caption{The impact of stochastic masking ratio on the performance of image editing tasks on the OmniEdit \cite{wei2024omniedit} dataset.}
  \resizebox{0.45\linewidth}{!}{
  \begin{tabular}{l ccc }
   \toprule
    Ratio&Edit Score $\uparrow$ &CLIP-T$\uparrow$& DINO $\uparrow$ \\ \hline
     
     0&0.742& 0.234& 0.767\\
 0.1&0.768& 0.235& \textbf{0.792}\\
 0.3&\textbf{0.779}& \textbf{0.236}& 0.790\\
     0.5&0.763& 0.235& 0.784\\ \toprule
  \end{tabular}
     }
  \label{tab:stochastic}
  \vspace{-2mm}
\end{table*}